\documentclass{article}
\usepackage[preprint]{log_2025}			

\usepackage{booktabs}						
\usepackage{multirow}						
\usepackage{amsfonts}						
\usepackage{graphicx}						
\usepackage{duckuments}						
\usepackage{float}
\usepackage{subcaption}
\usepackage{hyperref}
\usepackage{wrapfig}

\usepackage{tabularx,array,mathtools}

\newcolumntype{Y}{>{\raggedright\arraybackslash}X}

\usepackage[
     backend=biber,
     style=numeric-comp,
     backref=true,
     natbib=true]{biblatex}
\title{Structural Invariance Matters: Rethinking Graph Rewiring through Graph Metrics}

\author[Benoit et al.]{%
Alexandre Benoit\thanks{Equal contribution.}\\
University of Cambridge \\
\email{ab3149@cam.ac.uk}\And
Catherine Aitken\footnotemark[1]\\
University of Cambridge\\
\email{cpa32@cam.ac.uk}\And 
Yu He\\
Stanford University\\
\email{heyu@stanford.edu}
}

\begin{document}

\maketitle

\begin{abstract}
Graph rewiring has emerged as a key technique to alleviate over-squashing in Graph Neural Networks (GNNs) and Graph Transformers by modifying the graph topology to improve information flow. While effective, rewiring inherently alters the graph’s structure, raising the risk of distorting important topology-dependent signals. Yet, despite the growing use of rewiring, little is known about which structural properties must be preserved to ensure both performance gains and structural fidelity. In this work, we provide the first systematic analysis of how rewiring affects a range of graph structural metrics, and how these changes relate to downstream task performance. We study seven diverse rewiring strategies and correlate changes in local and global graph properties with node classification accuracy. Our results reveal a consistent pattern: successful rewiring methods tend to preserve local structure while allowing for flexibility in global connectivity. These findings offer new insights into the design of effective rewiring strategies, bridging the gap between graph theory and practical GNN optimization.
\end{abstract}

\section{Introduction}

Graph Neural Networks (GNNs) are a class of neural architectures designed to learn from graph-structured data \cite{sperduti, goller, gori, bruna2014spectralnetworkslocallyconnected, defferrard2017convolutionalneuralnetworksgraphs, kipf2017semisupervisedclassificationgraphconvolutional, NIPS1993_fc49306d}. They follow a message-passing paradigm \cite{Barceló_Kostylev_Monet_Pérez_Reutter_Silva_2019, pmlr-v70-gilmer17a}, where each node iteratively aggregates information from its neighbors. This simple yet powerful framework has enabled impactful applications in domains such as chemistry, social networks, and recommendation systems.

A central limitation of GNNs lies in their ability to propagate information across distant nodes. As the number of layers increases, a node’s receptive field expands exponentially, forcing large volumes of information to be compressed into fixed-size vectors. This phenomenon, known as \emph{over-squashing} \cite{topping2022understanding}, causes critical long-range signals to be lost. Over-squashing is closely tied to structural bottlenecks in the graph: when many long-range paths converge through narrow choke points, information is severely constrained. Recent work has linked this issue to topological properties such as curvature and effective resistance \cite{topping2022understanding}.

\paragraph{Problem Statement.} To mitigate bottlenecks and reduce over-squashing, recent research has explored \emph{graph rewiring}, which strategically modifies a graph’s structure to improve information flow. While these methods have demonstrated performance gains on benchmark tasks, their evaluation has focused almost exclusively on task accuracy. A critical but overlooked question is: \textbf{\emph{how much does rewiring preserve or distort the original structure?}} If rewiring alters key topological patterns—for example, by destroying characteristic connectivity motifs—the modified graph may no longer faithfully represent the underlying data \cite{barbero2024localityawaregraphrewiringgnns}.  This motivates the study of \emph{structural invariance}, the set of graph properties that remain unchanged under transformation. Structural invariance spans both local aspects (e.g., permutation invariance and equivariance) and global characteristics (e.g., Degree Distribution and connectivity patterns). Understanding which invariances matter most is especially important for architectures such as Graph Transformers, which rely on positional encodings and effectively operate on fully connected graphs, often at the expense of structural fidelity.  Despite the importance of preserving structural invariance, current rewiring methods are not evaluated in this regard. This creates a fundamental gap: although rewiring improves connectivity and mitigates over-squashing, we lack principled measures to assess which---and how much---of the graph’s inherent structure is preserved. Therefore, our goal is to compare rewiring methods not only in terms of GNN performance gains but also their topological fidelity, and to determine whether improved results come at the expense of fundamentally altering the graph’s nature. Addressing this question is critical for guiding the future design of graph rewiring strategies and positional encodings, by identifying which structural properties must be preserved to ensure faithfulness to the underlying data.






\paragraph{Contributions and Outline.}  
In this work, we introduce \textbf{GRASP} (Graph Rewiring Assessment of Structural Perturbation), a framework for systematically evaluating structural invariance in graph rewiring. Our key contributions are:  

\begin{itemize}
    \item We formalize structural invariance as a lens for evaluating graph rewiring and introduce a set of structural metrics tailored for this purpose (Section~\ref{sub:structural_metrics}).  
    \item We analyze how popular rewiring methods affect connectivity-related properties, revealing where and how structural fidelity is compromised (Section~\ref{sub:cbm}).  
    \item We study the sensitivity of different metrics to structural perturbations and connect these findings to downstream GNN performance, highlighting the trade-offs between accuracy gains and topological preservation (Section~\ref{sub:sim}).  
\end{itemize}

\section{Rewiring Methods}
\label{sec:related_work}


Although numerous graph rewiring methods have been proposed in recent years, classifying them into clear categories remains a challenge. Researchers commonly group these methods into two broad classes: spatial rewiring and spectral rewiring \cite{barbero2024localityawaregraphrewiringgnns}. Spatial rewiring focuses on altering the graph based on node proximity or positional information, such as connecting nodes within a certain number of hops. By using structural characteristics, spatial rewiring maintains much of the original graph’s relevant structural information while adding edges to reduce over-squashing. Spectral rewiring centers on modifying connectivity to optimize global spectral properties of the graph (such as the spectral gap). Methods in this category tend to add edges between distant nodes in order to improve long-range information flow, thus reducing over-squashing without necessarily preserving local structure to the same extent as spatial methods. The most prominent rewiring methods used in this paper are summarized below in chronological order and more details are in Appendix~\ref{appendix:rewiring methods}.

Concretely, we evaluate: DiffWire \cite{diffwire}, which differentiably optimizes commute times and $\lambda_2$ within the GNN; SDRF \cite{sdrf}, which rewires along edges with negative (Balanced Forman) curvature; GTR \cite{gtr}, which greedily minimizes $R_G$; BORF \cite{borf}, which removes highly positively curved edges (over-smoothing) and adds edges in negatively curved regions (over-squashing); FOSR \cite{fosr}, which selects edges predicted to maximize $\lambda_2$; and LASER \cite{laser}, which incrementally adds locality-preserving $n$-hop edges where few short walks exist. Full algorithmic details are deferred to the appendix.

\section{GRASP: Graph Rewiring Assessment of Structural Perturbation}

\label{sub:structural_metrics}

We evaluate structural invariance by quantifying changes in a range of structural metrics under different rewiring methods and relating these changes to performance across four benchmark datasets.

\begin{table*}[t]
\centering
\footnotesize
\setlength{\tabcolsep}{6pt}
\renewcommand{\arraystretch}{1.2}
\begin{tabularx}{\textwidth}{@{} l l Y @{}}
\toprule
\textbf{Metric} & \textbf{Type} & \textbf{Formula} \\
\midrule

Diameter & Connectivity, Spatial &
\(\displaystyle \max_{u,v \in \mathrm{CC}(G)} d(u,v)\) \\

Effective graph resistance & Connectivity, Spectral &
\(\begin{aligned}[t]
R_G &= \sum_{1\le i<j\le n} R_{ij},\\
R_{ab} &= \frac{v_a - v_b}{I}
\end{aligned}\) \\

Modularity & Structural, Spatial &
\(\begin{aligned}[t]
Q &= \sum_{c=1}^{n}\!\left[\frac{L_c}{m}
-\gamma\!\left(\frac{k_c}{2m}\right)^{\!2}\right]
\end{aligned}\) \\

Degree assortativity & Structural, Spatial &
\(\begin{aligned}[t]
r &= \frac{\sum_{x,y} xy\,\bigl(e_{xy}-a_x b_y\bigr)}
{\sigma_a \sigma_b}
\end{aligned}\) \\

Global clustering coefficient & Structural, Spatial &
\(\begin{aligned}[t]
C &= \frac{1}{n}\sum_{v} c_v,\\
c_u &= \frac{2T(u)}{\deg(u)\,(\deg(u)-1)}
\end{aligned}\) \\

Spectral gap & Connectivity, Spectral &
\(\lambda_2(L)\) \\

Forman curvature  & Connectivity, Spatial &
\(\begin{aligned}[t]
F_{\mathrm{full}}(e)
&= w_e\Bigg(
\frac{w_{v_1}}{w_e\, w_{e v_1}}
+
\frac{w_{v_2}}{w_e\, w_{e v_2}}
\\[-2pt]&\quad
-\!\!\sum_{\substack{e_{v_1}\sim e\\ e_{v_2}\sim e}}
\!\!\left[
\frac{w_{v_1}}{\sqrt{w_e\, w_{e v_1}}}
+
\frac{w_{v_2}}{\sqrt{w_e\, w_{e v_2}}}
\right]
\Bigg)
\end{aligned}\) \\

Average betweenness centrality & Structural, Spatial &
\(\begin{aligned}[t]
\mathrm{ABC} &= \frac{1}{|G|}\sum_{n} c_n,\\
c_B(v) &= \sum_{s,t\in V}
\frac{\sigma(s,t\,|\,v)}{\sigma(s,t)}
\end{aligned}\) \\

Jaccard similarity & Similarity &
\(\displaystyle J(A,B)=\frac{|A\cap B|}{|A\cup B|}\) \\

Laplacian spectrum distance & Similarity, Spectral &
\(\begin{aligned}[t]
d_{\mathrm{Lap}}(G,G')
&= \Bigg(\sum_{i=1}^{n}
\bigl|\lambda_i-\lambda'_i\bigr|^{p}\Bigg)^{\!1/p}
\end{aligned}\) \\

Adjacency spectral norm diff. & Similarity &
\(\displaystyle d_{\mathrm{Adj}}(G,G')=\|A-A'\|_{2}\) \\

Degree Distribution distance & Similarity &
\(\begin{aligned}[t]
d_{\mathrm{Deg}}(G,G')
&= W_1(P,Q)\\
&= \inf_{\gamma\in\Gamma(P,Q)}
\ \mathbb{E}_{(x,y)\sim\gamma}\bigl[\,|x-y|\,\bigr]
\end{aligned}\) \\

Shortest-path length dist. & Similarity &
\(\displaystyle W_1\!\big(P_{\mathrm{sp}}(G),\,P_{\mathrm{sp}}(G')\big)\) \\

\bottomrule
\end{tabularx}
\caption{Summary of structural, connectivity, and similarity metrics used in our framework.}
\label{tab:metrics-compact}
\end{table*}

\subsection{Structural Metrics}

As our aim is to better understand structural invariance through rewiring and its impact on downstream performance, we choose metrics to cover both overall connectivity and inherent graph topological information. Thus, metrics can be divided into two groups.
\begin{itemize}
    \item Those that inherently must change to improve information flow within a graph, such as those relating to the existence of bottlenecks, and so are expected to change.
    \item Those that indicate structural context within a graph.
\end{itemize} 
We use these metrics as an indicator of a graph's global patterns to portray the overall impact of these rewiring methods. To do this, metrics that are calculated on a single node are averaged across all nodes in a graph. We choose more spatial metrics than spectral metrics to focus on structural properties, but still consider spectral properties through the use of the spectral gap and effective resistance metrics. The detailed description of these metrics can be found in Appendix~\ref{appendix:metrics equations}.

\label{sec:strumetr}

We characterize graph structure using complementary metrics: diameter, capturing global reachability via the longest shortest path \cite{doi:10.1073/pnas.252631999}; effective graph resistance, indicating ease of flow where lower values reflect many parallel routes and fewer bottlenecks \cite{ELLENS20112491}; modularity, quantifying community strength as dense within-group and sparse between-group connectivity \cite{Clauset_2004,doi:10.1073/pnas.0601602103}; degree assortativity, measuring preference for like-degree attachment (positive for hub–hub, negative for hub–periphery) \cite{Newman_2002,Newman_2003}; the global clustering coefficient, summarizing triangle-based local cohesion, with low values highlighting tree-like, bottleneck-prone regions \cite{10.1093/oso/9780198805090.003.0007}; the spectral gap (algebraic connectivity), a proxy for mixing speed and robustness where larger gaps imply stronger connectivity \cite{10.1093/imrn/rnz077}; Forman curvature, a local indicator of geodesic convergence/divergence that flags structural bottlenecks linked to over-squashing in GNNs \cite{topping2022understanding,forman,ni2019community,bober2022rewiringnetworksgraphneural}; and average betweenness centrality, reflecting how concentrated shortest-path traffic is across nodes  \cite{10.1093/oso/9780198805090.003.0007}.

\subsection{Metrics Calculation Strategy}

We calculate the mean and standard deviation of each of these metrics before and after rewiring foreach dataset of the \texttt{TUDataset} group. Our work excludes the COLLAB and REDDIT-BINARY datasets (Table~\ref{tab:dataset_characteristics}) due to CPU memory constraints, as their sizes prevented the practical implementation of some rewiring methods, such as Diffwire, that required learning dataset-specific parameters. 

 Following this initial collection, we create a framework to dynamically implement the various rewiring techniques on each of the chosen datasets. This excludes the Diffwire method, as it requires significant compute to learn the commute-time embeddings for rewiring. We instead implemented Diffwire following the CT-Layer tutorial provided by the authors of the original Diffwire paper \cite{arnaizrodriguez2022diffwireinductivegraphrewiring}. The rewired datasets are then used to recompute and compare the structural metrics and respective performances using GNN accuracy results in \cite{barbero2024localityawaregraphrewiringgnns}.

\subsection{Similarity Metrics Collection}
For further analysis, we calculate five similarity metrics between the original and rewired datasets to analyze structural invariance through rewiring on the \texttt{MUTAG}, \texttt{ENZYMES}, and \texttt{PROTEINS} datasets. We constrain our analysis to three rewiring methods for brevity: the highest-performing, middle-performing (fourth-highest) and lowest-performing methods. 

To quantify how rewiring alters graph structure, we employ complementary similarity and distance measures. Jaccard similarity over edge sets captures the direct overlap of edges between the original and rewired graphs, emphasizing literal preservation or replacement of connections \cite{9926326}. The Laplacian spectrum distance compares their sorted Laplacian eigenvalues, sensitively reflecting global structural changes that affect connectivity, diffusion, and community signals \cite{PATANE201968}. The spectral norm of the adjacency difference focuses on the largest singular deviation between adjacency matrices, highlighting worst-case shifts in connectivity patterns (e.g., a few strong edges added or removed) \cite{gervens2022graphsimilaritybasedmatrix}. The degree-distribution distance uses the Wasserstein (W1) metric to measure how much “mass” must move to transform one degree profile into the other, revealing shifts in hub–periphery balance and overall heterogeneity \cite{Panaretos_2019}. Finally, the W1 distance between shortest-path length distributions evaluates changes in reachability and routing efficiency across the graph after rewiring \cite{10.1007/978-3-642-17316-5_32}.

\section{Results and Analysis}
\label{sec:results}

The aforementioned metrics are divided into two sections: 
\begin{itemize}
    \item \textbf{Connectivity-based metrics}: Metrics that are expected to change when the graph's connectivity is modified, reflecting improvements in information flow due to solving bottleneck issues.
    \item \textbf{Structural information metrics}: Metrics that capture inherent structural patterns or properties of the graph, providing insight into its local and global organization.
\end{itemize}

Metrics values are ordered from highest to lowest-performing rewiring method \cite{barbero2024localityawaregraphrewiringgnns}, with GCN classification accuracy shown as overlaid points on the bar chart for each method.

On the right-hand side, we display the percentage change of the mean of each metric from the unrewired dataset to the rewired datasets. 

The size of dataset \texttt{REDDIT-BINARY} prevented the practical implementation of all rewiring techniques save for GTR and BORF. These values are included for additional context, but are not used in analysis. 

\subsection{Connectivity-based Metrics}
\label{sub:cbm}
\begin{figure}[t]
    \centering
    \begin{subfigure}[b]{0.49\textwidth}
        \centering
        \includegraphics[width=\linewidth]{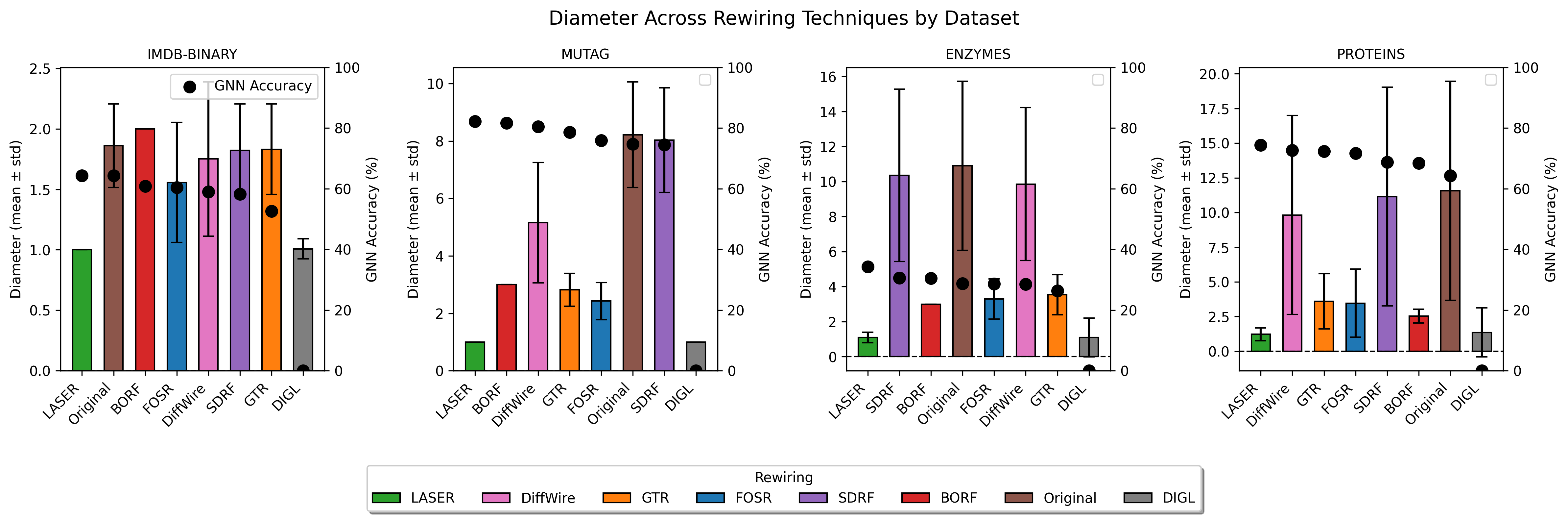}
        \caption{Diameter Values}
        \label{fig:diam_true}
    \end{subfigure}
    \hfill
    \begin{subfigure}[b]{0.49\textwidth}
        \centering
        \includegraphics[width=\linewidth]{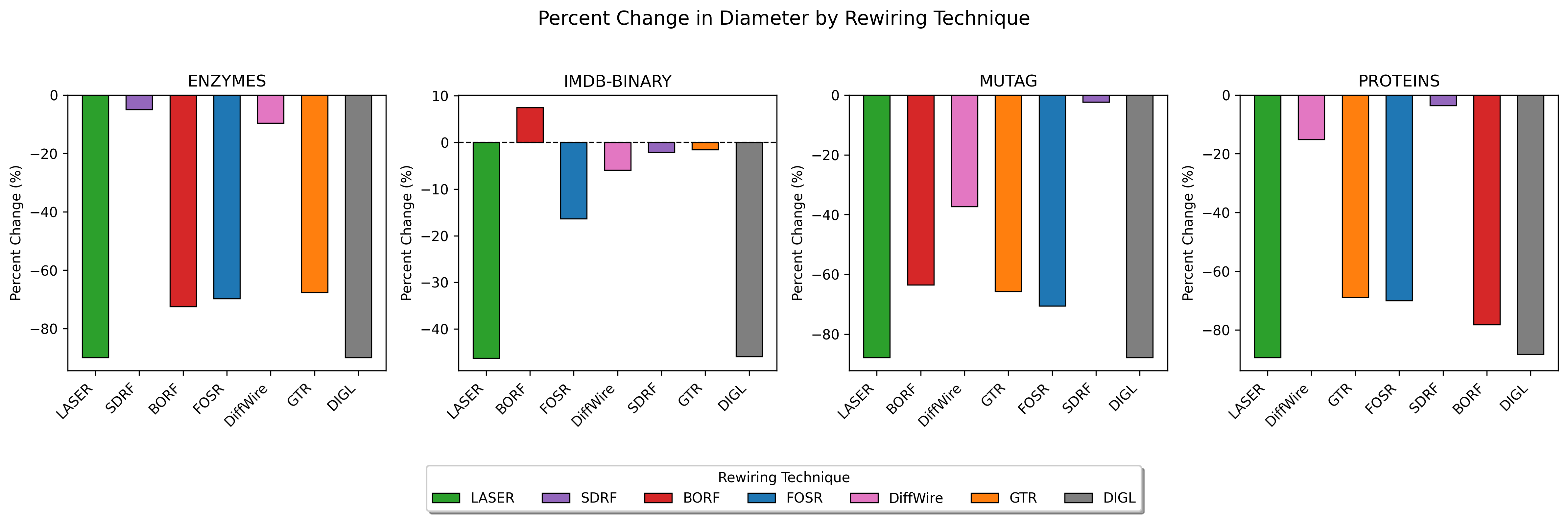}
        \caption{Percentage Change in Diameter}
        \label{fig:diam_perc}
    \end{subfigure}

    \begin{subfigure}[b]{0.49\textwidth}
        \includegraphics[width=\linewidth]{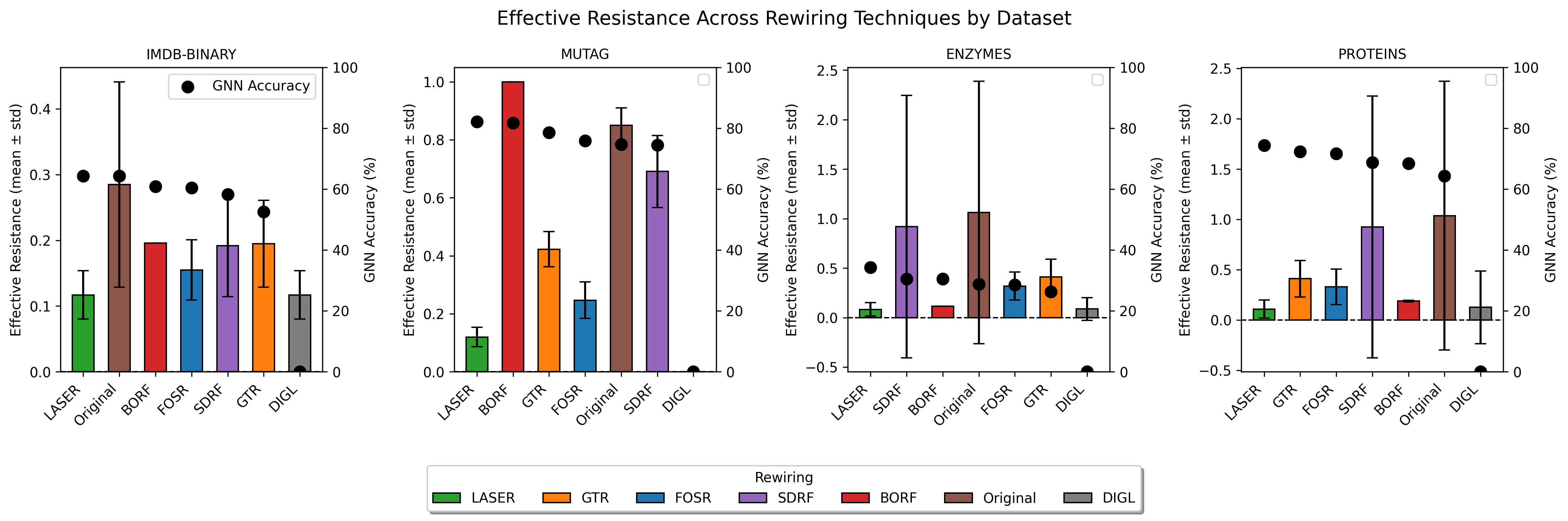}
        \caption{Effective Resistance Values}
        \label{fig:er_true}
    \end{subfigure}
    \hfill
    \begin{subfigure}[b]{0.49\textwidth}
        \includegraphics[width=\linewidth]{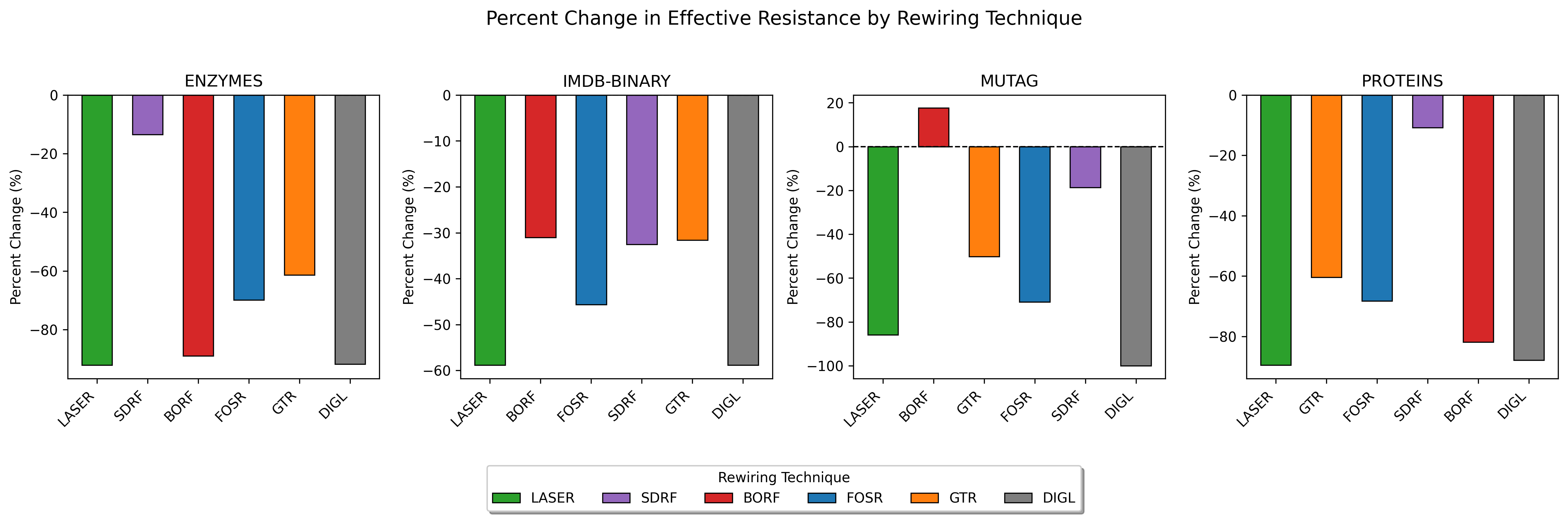}
        \caption{Percentage Change in Effective Resistance}
        \label{fig:er_pc}
    \end{subfigure}
    \begin{subfigure}[b]{0.49\textwidth}
        \includegraphics[width=\linewidth]{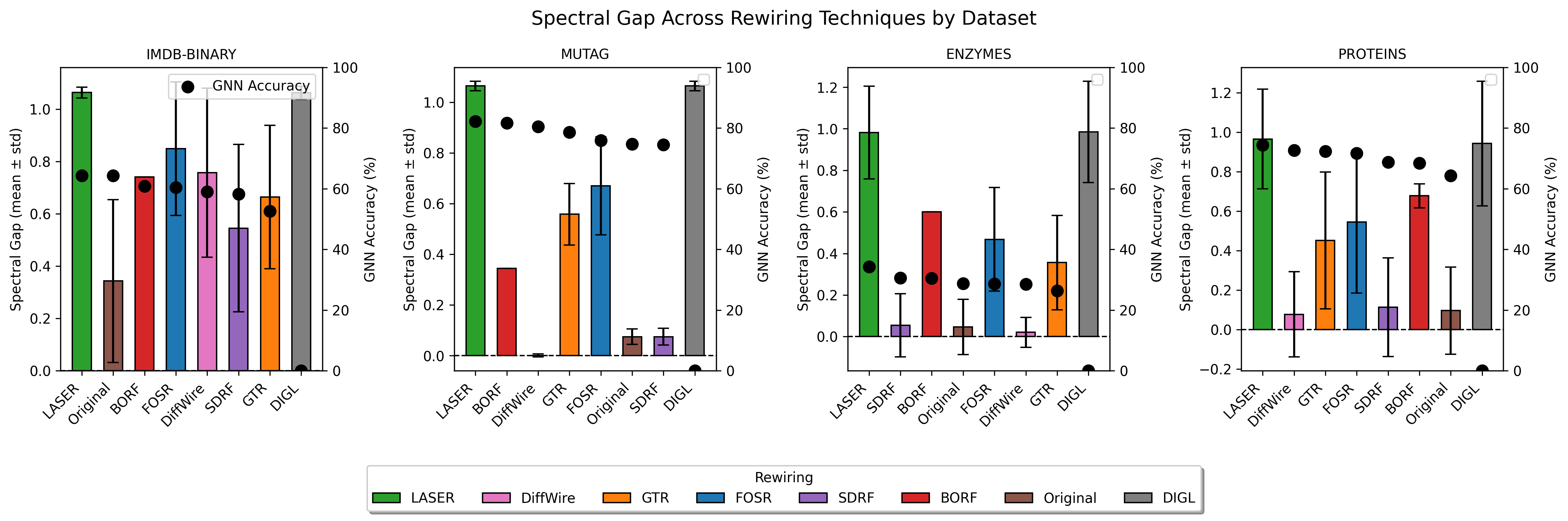}
        \caption{Spectral Gap Values}
        \label{fig:sg_true}
    \end{subfigure}
    \hfill
    \begin{subfigure}[b]{0.49\textwidth}
        \includegraphics[width=\linewidth]{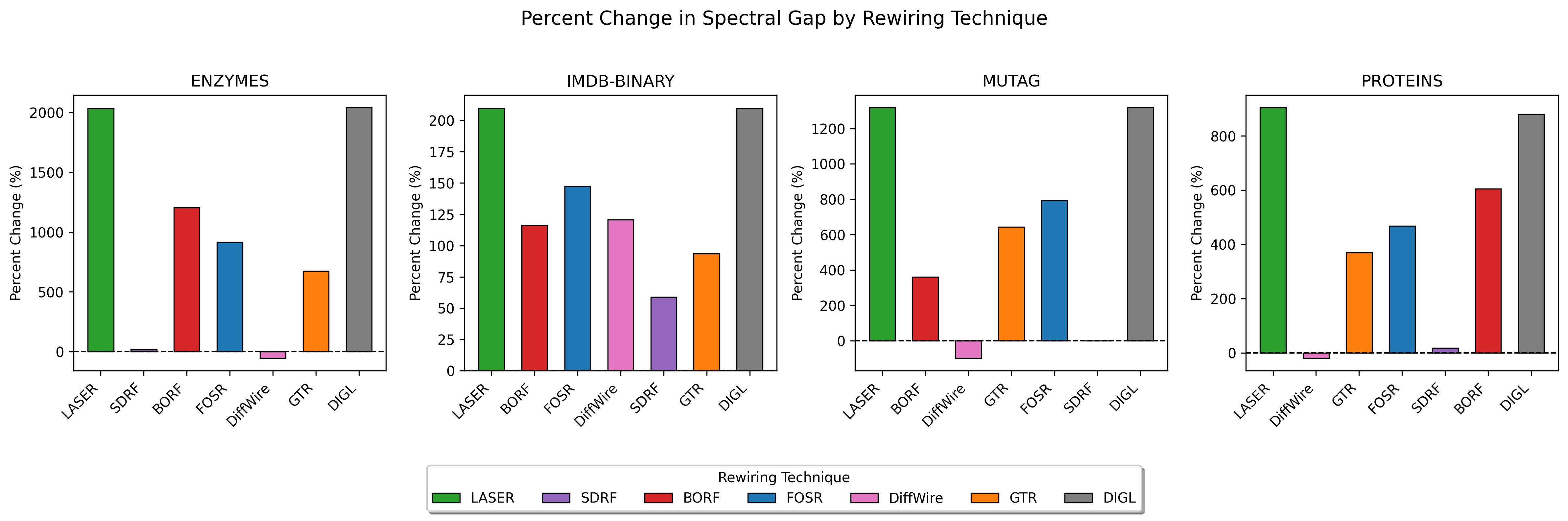}
        \caption{Percentage Change in Spectral Gap}
        \label{fig:sg_pc}
    \end{subfigure}
    \begin{subfigure}[b]{0.49\textwidth}
        \includegraphics[width=\linewidth]{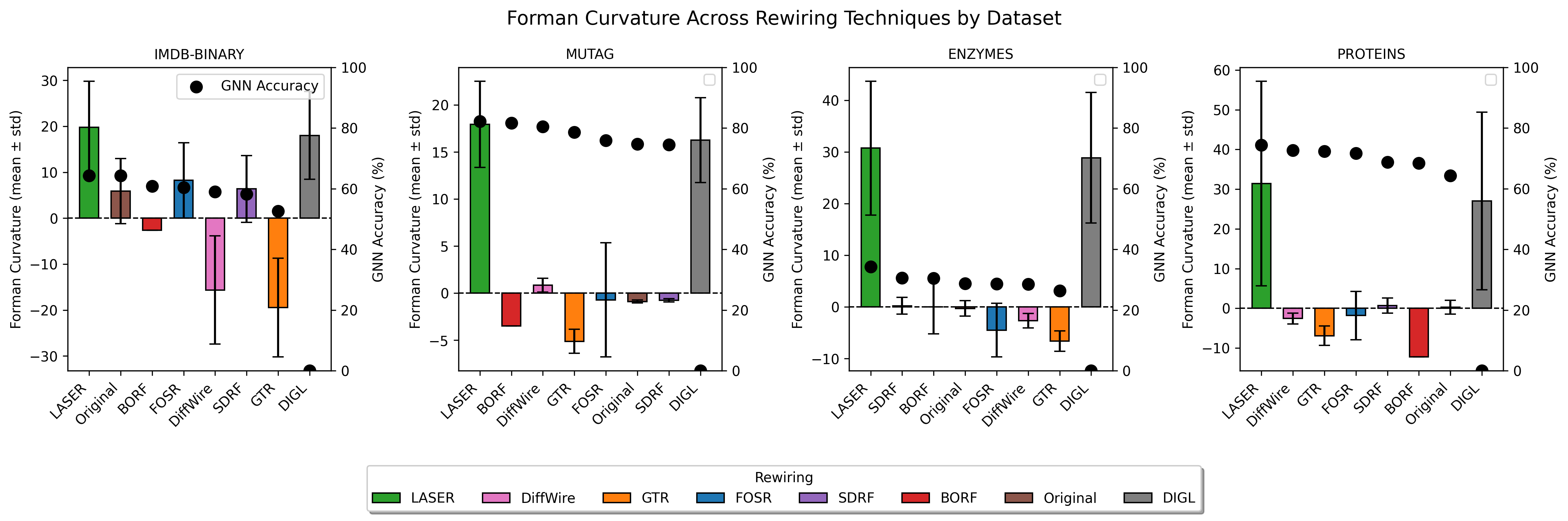}
        \caption{Forman Curvature Values}
        \label{fig:forman_Val}
    \end{subfigure}
    \hfill
    \begin{subfigure}[b]{0.49\textwidth}
        \includegraphics[width=\linewidth]{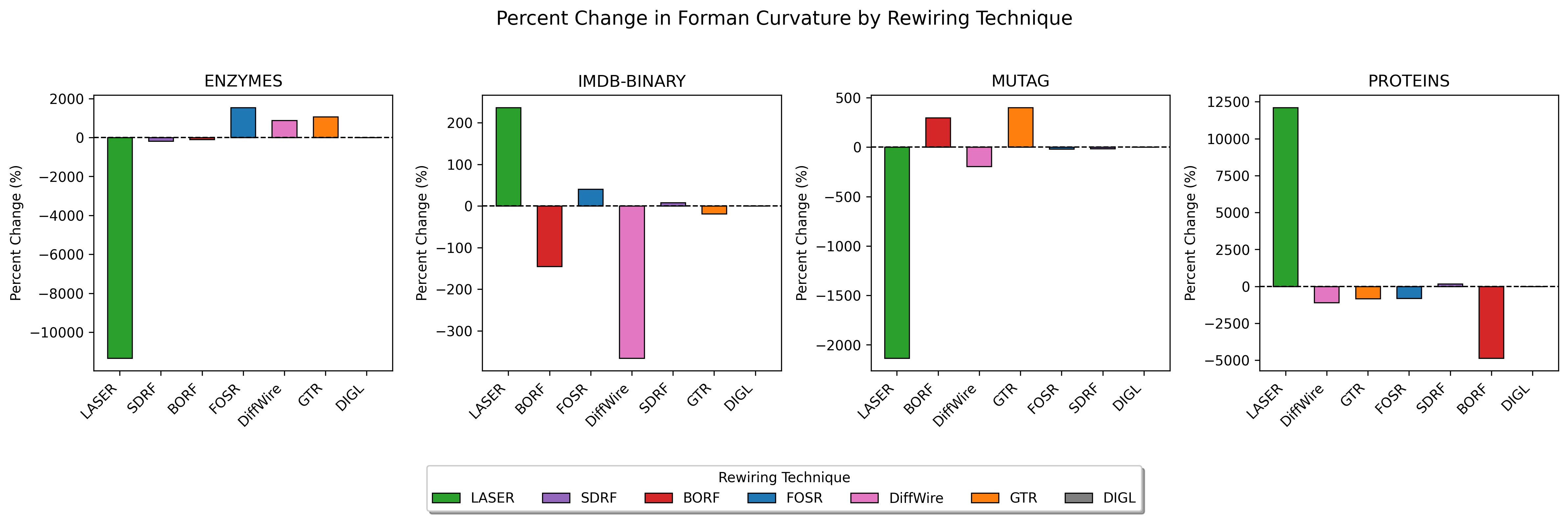}
        \caption{Percentage Change in Forman Curvature}
        \label{fig:pcforman}
    \end{subfigure}
    \caption{Connectivity metrics over all datasets and rewiring methods with corresponding percentage change and GNN accuracy calculated in \cite{laser}. The percentage change due to rewiring is calculated as the difference between the rewired and original graphs, divided by the original graph.}
    \label{fig:metricandpercentchange}
\end{figure}

\subsubsection{Diameter}

Diameter almost always decreased through rewiring, especially in the highest performing rewiring methods. \textbf{LASER} resulted in the greatest decrease in diameter by percentage over all the rewiring techniques (Figure ~\ref{fig:diam_perc}). GNNs rely on propagating information across edges, so reducing diameter facilitates message-passing to all nodes within fewer hops. This is an important improvement as GNNs have a fixed number of layers.

One outlier is the \texttt{ENZYMES} dataset, in which diameter remains relatively unchanged after the second-highest performing rewiring method, \textbf{SDRF} (Figure ~\ref{fig:diam_true}). This may indicate node classification for \texttt{ENZYMES} relies more on local rather than global information. 

\subsubsection{Effective Resistance}
Effective resistance (Figure~\ref{fig:er_true} \&~\ref{fig:er_pc}) also generally decreases through rewirings, especially in \textbf{LASER}. The largest decrease in effective resistance correlates with the greatest performance, suggesting effective resistance could be an important indicator of improved connectivity. As effective resistance is essentially a graph's resistance to information flow, reducing this metric facilitates message-passing. 

However, this is not the case for all datasets and rewiring methods. \textbf{BORF} is the second-highest performing method for the \texttt{MUTAG} dataset, yet increases effective resistance. \textbf{BORF} is the only rewiring method that aggressively removes edges (Figure ~\ref{tab:percentage_change_edgesaddedremoved}), which explains the increase in effective resistance. Considering this, effective resistance does not need to decrease in cases where local community structure is more critical than global structure, which is potentially the case with \texttt{MUTAG}.

\textbf{Diffwire} is removed from the effective resistance plots, as it significantly increased effective resistance for all datasets and so was not comparable to the other rewiring techniques. 

\subsubsection{Spectral Gap}
\label{sec:spec_gap}
The spectral gap (Figures~\ref{fig:sg_true} \&~\ref{fig:sg_pc}) generally increases for the highest performing rewiring techniques, with \textbf{LASER} increasing this metric the most. As a large spectral gap can be indicative of improved connectivity, it is understandable that an increase in this metric corresponds to better performance.

The spectral gap is also called the algebraic connectivity \cite{abdi2020regulargraphsminimumspectral} and can be seen as the difficulty required to divide the graph into distinct parts. Adding long-range connections through rewiring would make this division more difficult, explaining the increase in spectral gap. Datasets such as \texttt{ENZYMES}  and \texttt{MUTAG}, for which effective resistance only dramatically increases with \textbf{LASER}, may originally be difficult to divide. In this case, improving connectivity by increasing the spectral gap may not result in as notable of an performance improvement.

\subsubsection{Forman Curvature}

Figure~\ref{fig:forman_Val} illustrates that techniques such as \textbf{LASER} and \textbf{DIGL} often increase curvature substantially, while approaches like \textbf{SDRF} or \textbf{DiffWire} reduce curvature below the original levels. This variation stems from each method's design priorities - whether they emphasize local/global connectivity, or rely on spectral/spatial strategies. Large positive jumps in curvature typically correspond to reduced \textbf{over-squashing}, suggesting successful mitigation of bottlenecks, by reinforcing weakly connected regions (Appendix~\ref{appendix:curvature} - Figure~\ref{fig:curvature_dist}).

In datasets like \texttt{IMDB-BINARY} and \texttt{PROTEINS}, which already exhibit high curvature (Figure~\ref{fig:pcforman}), further rewiring yields diminishing returns. Additionally, higher curvature doesn't always lead to better performance; especially when local adjacency is already strong or node features dominate. In such cases, heavily rewiring may disrupt structural invariance and harm downstream accuracy. Curvature trends are most informative when viewed alongside GNN accuracy: if both rise, the rewiring likely resolved meaningful bottlenecks, but if curvature increases while accuracy plateaus or drops, the graph's useful structure may have been compromised.


Curvature-based rewiring is often more \textbf{surgical}, preserving the original topology more effectively (\textbf{LASER}) than random-walk methods. While \textbf{LASER} tends to retain local features, it can introduce excessive edges, leading to dense graphs that obscure critical relationships. Hence, moderate curvature increases often strike the best balance between improving connectivity and preserving structural integrity - a trend also reflected in metrics like assortativity (Section:\ref{subsub:assor}), where maintaining \emph{just enough} locality correlates with stronger performance.


\subsection{Structural Information Metrics}
\label{sub:sim}
\begin{figure}[h]
    \centering
    
    \begin{subfigure}[b]{0.49\textwidth}
        \includegraphics[width=\linewidth]{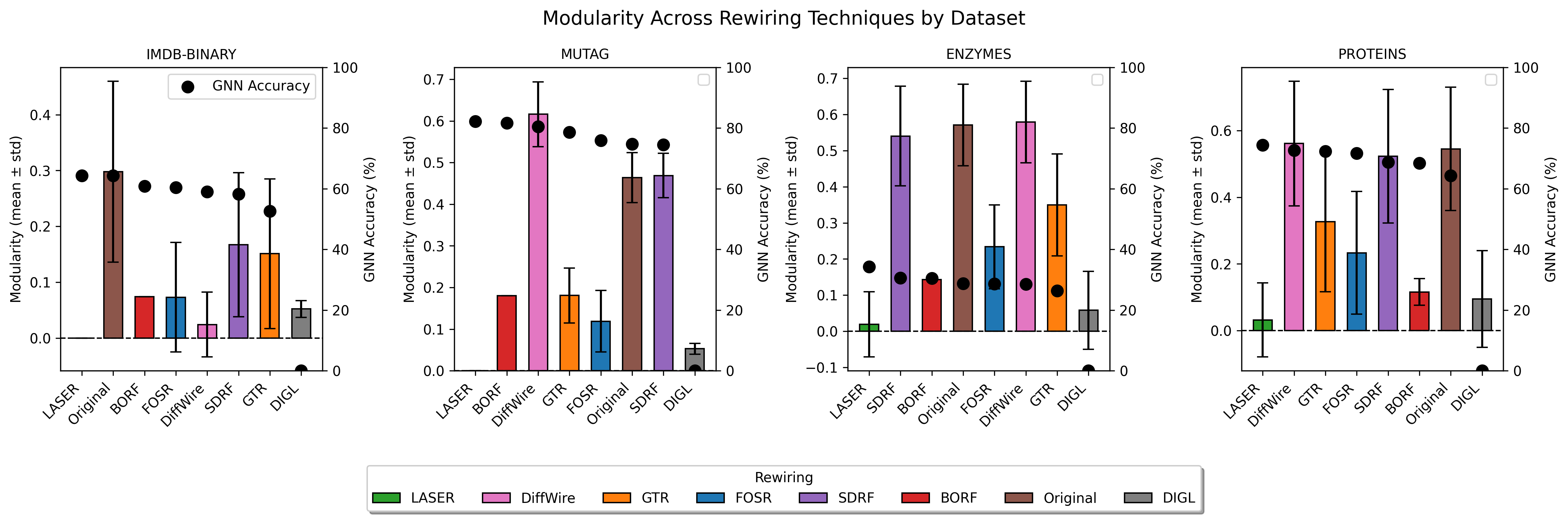}
        \caption{Modularity Values}
        \label{fig:mod_true}
    \end{subfigure}
    \hfill
    \begin{subfigure}[b]{0.49\textwidth}
        \includegraphics[width=\linewidth]{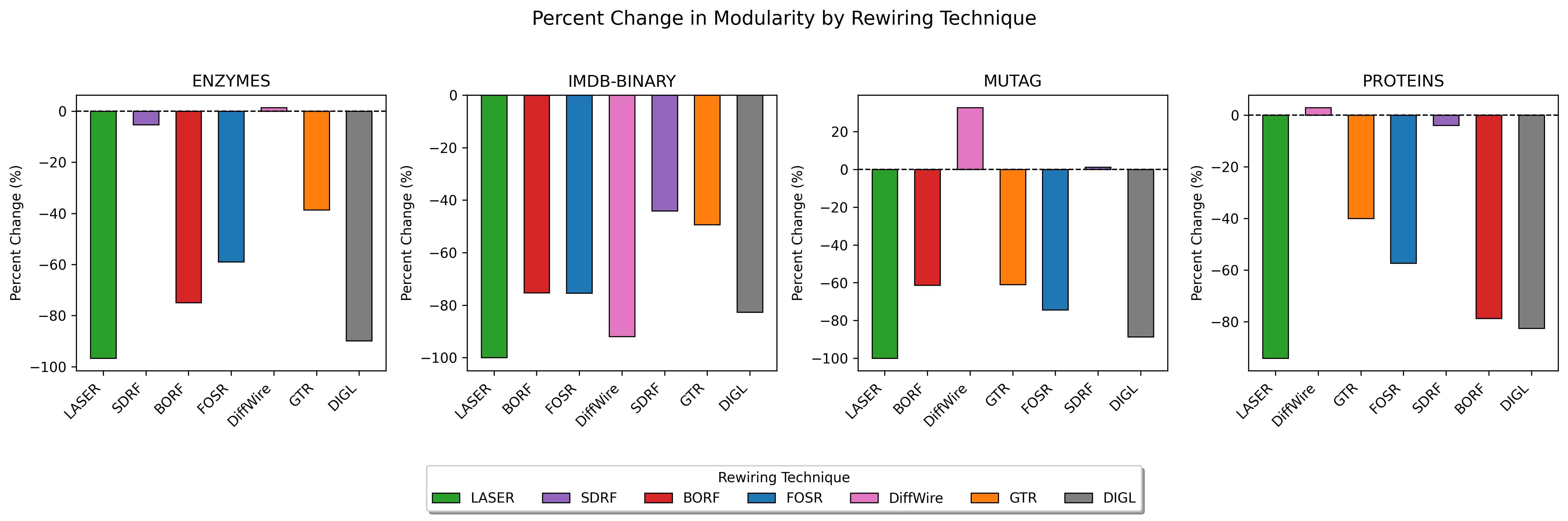}
        \caption{Percentage Change in Modularity}
        \label{fig:mod_pc}
    \end{subfigure}
    \begin{subfigure}[b]{0.49\textwidth}
        \includegraphics[width=\linewidth]{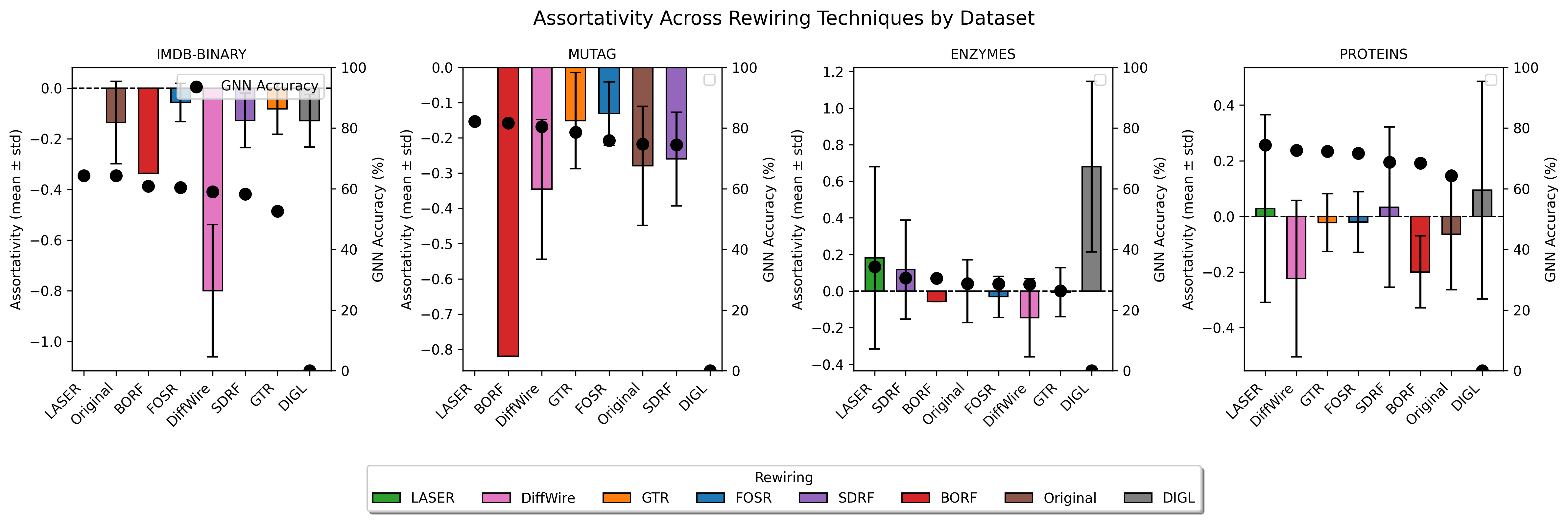}
        \caption{Assortativity}
        \label{fig:assor_true}
    \end{subfigure}
    \hfill
    \begin{subfigure}[b]{0.49\textwidth}
        \includegraphics[width=\linewidth]{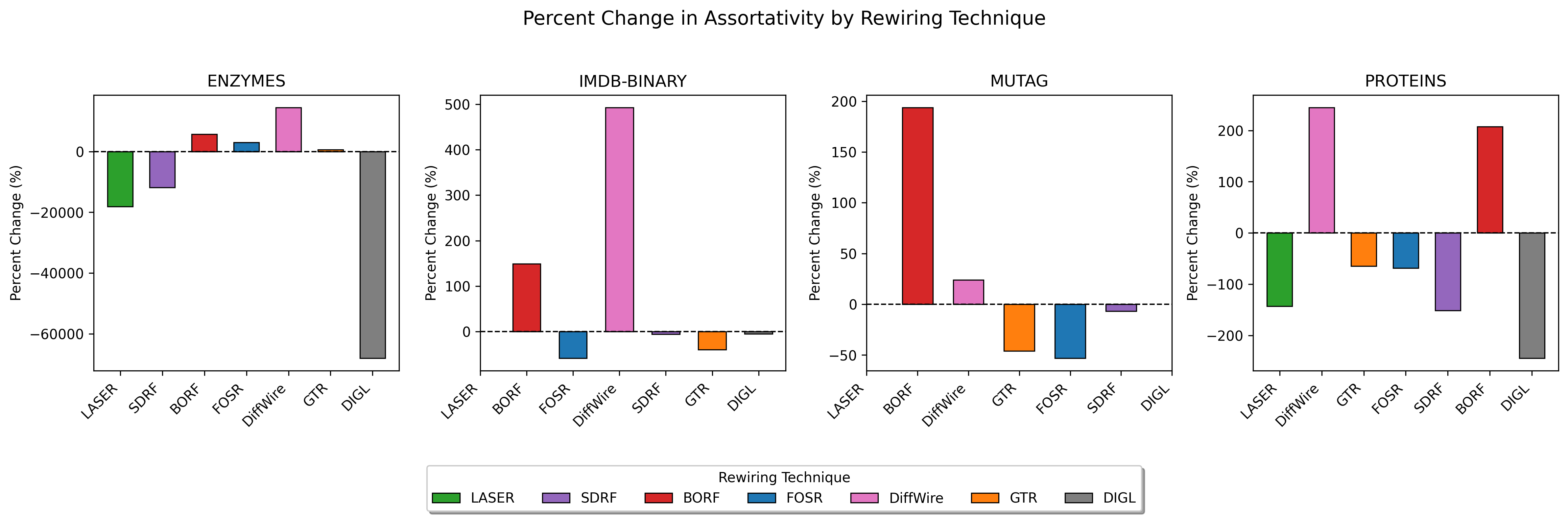}
        \caption{Percentage Change in Assortativity}
        \label{fig:assor_pc}
    \end{subfigure}
    \begin{subfigure}[b]{0.49\textwidth}
        \includegraphics[width=\linewidth]{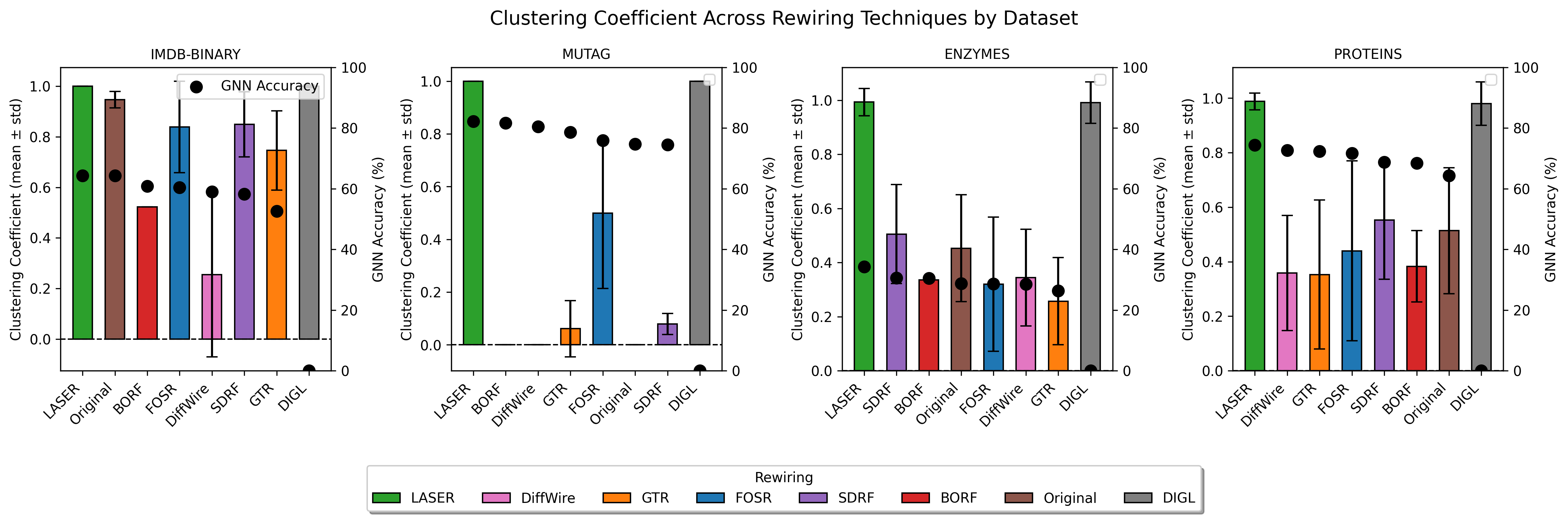}
        \caption{Clustering Coefficient Values}
        \label{fig:cc_true}
    \end{subfigure}
    \hfill
    \begin{subfigure}[b]{0.49\textwidth}
        \includegraphics[width=\linewidth]{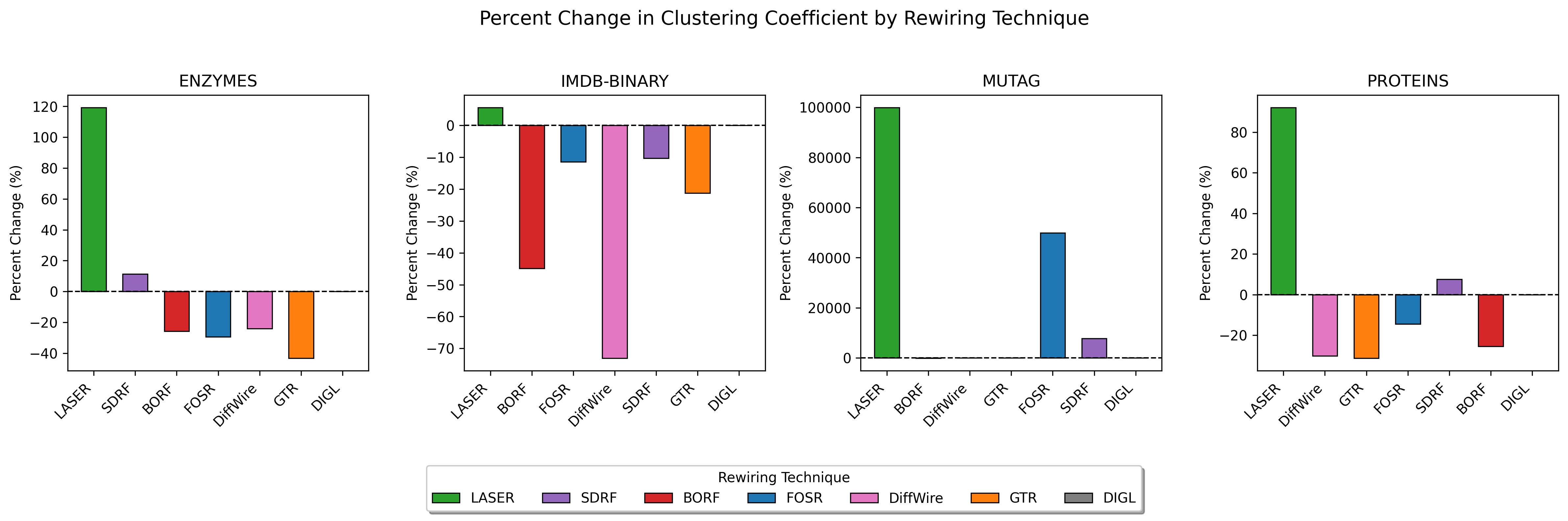}
        \caption{Percentage Change in Clustering Coefficient}
        \label{fig:ccpc}
    \end{subfigure}
    \begin{subfigure}[b]{0.49\textwidth}
        \includegraphics[width=\linewidth]{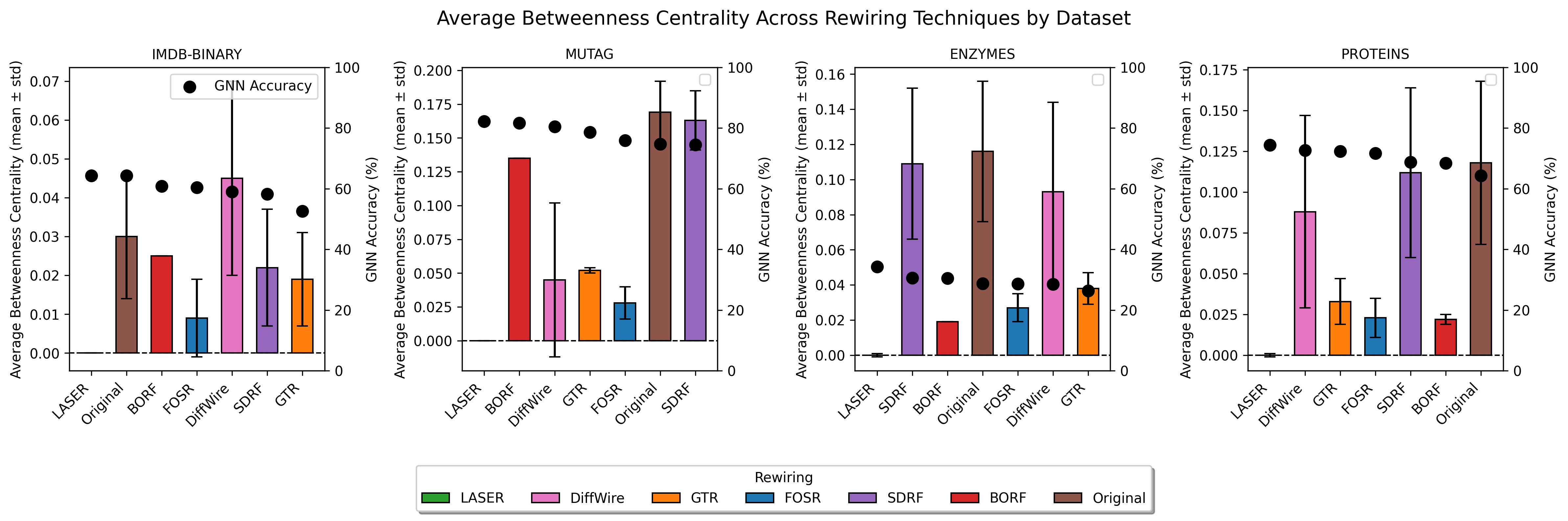}
        \caption{Average Betweenness Centrality}
        \label{fig:abc_true}
    \end{subfigure}
    \hfill
    \begin{subfigure}[b]{0.49\textwidth}
        \includegraphics[width=\linewidth]{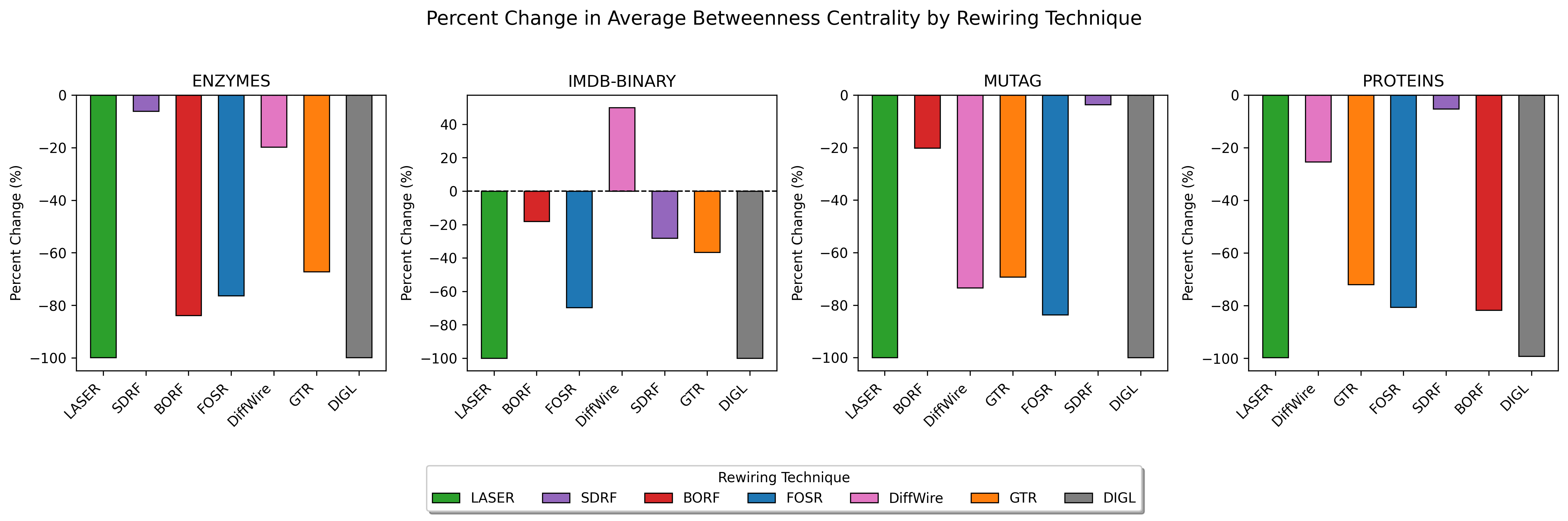}
        \caption{Percentage Change in Average Betweenness Centrality}
        \label{fig:abc_pc}
    \end{subfigure}

    \caption{Structural information metrics over all datasets and rewiring methods with corresponding percentage change and GNN accuracy calculated in \cite{laser}. The percentage change is calculated as the difference between the rewired and original graphs, divided by the original graph.}
    \label{fig:comm_metricandpercentchange}
\end{figure}

\subsubsection{Modularity}
Modularity (Figure~\ref{fig:mod_true} \& ~\ref{fig:mod_pc}) generally decreases across all rewiring techniques and datasets, save for \textbf{Diffwire} applied to the \texttt{MUTAG} dataset. \textbf{LASER} rewiring results in fully-connected graphs for the datasets with smaller graph sizes, \texttt{MUTAG} and \texttt{IMDB-BINARY}. Fully connected graphs do not maintain a distinct community structure, explaining the decrease in modularity to near-zero values. This decrease in clear community structure could indicate an increase in interconnections between previously distinct groups.

The main exception is \textbf{SDRF}'s high performance on the \texttt{ENZYMES} dataset while not appreciably decreasing its modularity. This is possibly due to the same reasons as outlined in Section ~\ref{sec:spec_gap} for the \texttt{ENZYMES} dataset. The benefit of decreasing modularity through rewiring diminishes for datasets that are originally strongly connected and lack clear community structure.

\subsubsection{Assortativity}
\label{subsub:assor}
Assortativity is the first metric in which the successful rewiring techniques do not significantly change the average value. \textbf{LASER} and \textbf{SDRF} maintain assortativity invariance relative to other rewiring techniques like \textbf{Diffwire} and \textbf{BORF}. \textbf{Diffwire} only performs within the top three rewiring methods for the dataset \texttt{MUTAG}, which is the dataset in which it alters the assortativity least. These patterns suggest a graph's degree assortativity is a characteristic that is preserved by successful rewiring methods. 

An important aspect to note is that \textbf{LASER} results in undefined average degree assortativity for the \texttt{MUTAG} and \texttt{IMDB-BINARY} datasets since the rewired graphs are fully-connected. 

\subsubsection{Clustering Coefficient}
The clustering coefficient (Figure~\ref{fig:cc_true} \& ~\ref{fig:ccpc}) generally remains invariant or decreases compared to the original datasets. This could be indicative of long-range connections, diminishing the importance of local community structure and increasing global communication. The tendency for rewiring techniques to maintain or reduce the average clustering coefficient suggests that supporting global message-passing is important for success, but not to the extent of fully displacing local community composition.

The main outlier is \textbf{LASER}, which consistently modifies the graphs within a dataset to achieve an average clustering coefficient around 1.00; a noticeable increase from the unrewired average clustering coefficient for all datasets except \texttt{IMDB-BINARY}. This is due to \textbf{LASER}'s tendency to create fully-connected graphs. 

\subsubsection{Average Betweenness Centrality}
The average betweenness centrality (ABC) (Figure~\ref{fig:abc_true} \& ~\ref{fig:abc_pc}) generally decreases among successful rewiring techniques, although by varying magnitudes. \textbf{LASER} results in an ABC value of zero as it creates a fully-connected graph. 

A decrease in ABC indicates greater general connectivity as there are fewer nodes that act as "central" nodes along paths, as is the case in bottlenecks. More alternate short-paths exist in the rewired graphs, reducing the need for information to flow through just a few select nodes. 

\subsection{Similarity Metrics}

\begin{figure}[h]
            \centering
            \begin{subfigure}[b]{0.45\textwidth}
                \includegraphics[width=\textwidth]{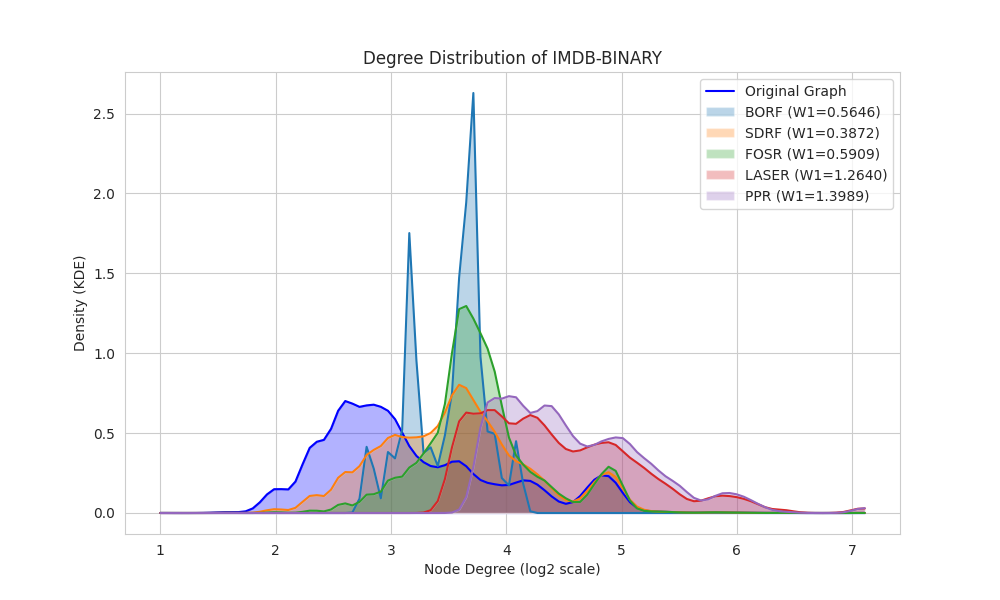}
                \caption{Original vs Rewired Degree Distribution for IMDB-BINARY}
                \label{fig:URDDI}
            \end{subfigure}
            \hfill
            \begin{subfigure}[b]{0.45\textwidth}
                \includegraphics[width=\textwidth]{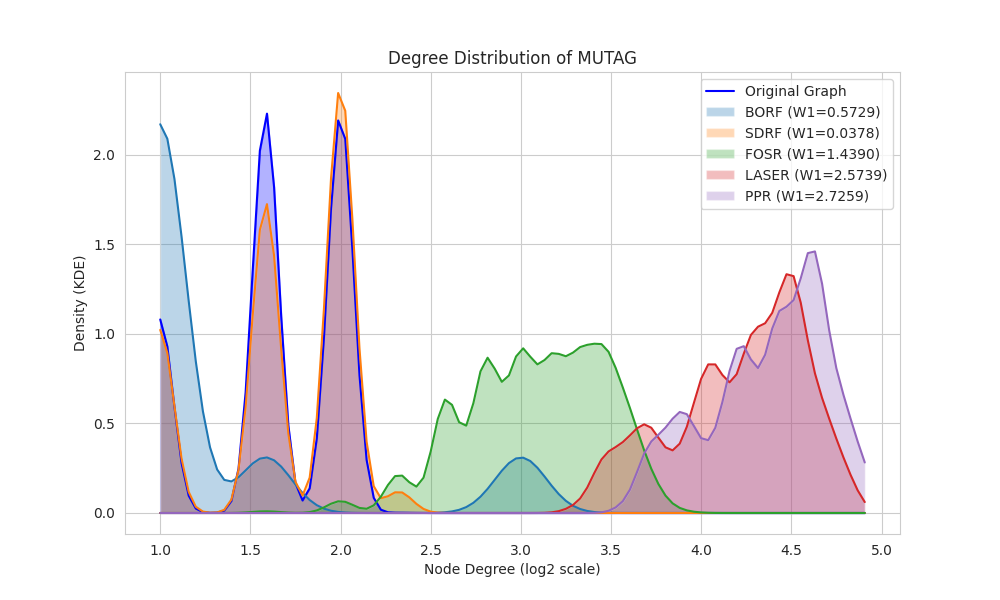}
                \caption{Original vs Rewired Degree Distribution for MUTAG}
                \label{fig:URDDM}
            \end{subfigure}

            \begin{subfigure}[b]{0.45\textwidth}
                \includegraphics[width=\textwidth]{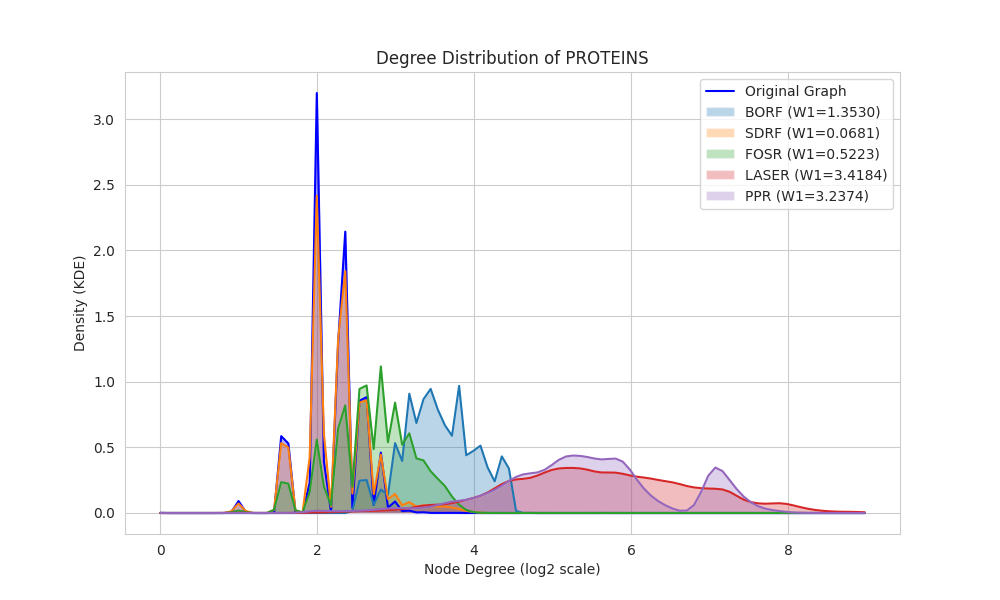}
                \caption{Original vs Rewired Degree Distribution for PROTEINS}
                \label{fig:URDDP}
            \end{subfigure}
            \hfill
            \begin{subfigure}[b]{0.45\textwidth}
                \includegraphics[width=\textwidth]{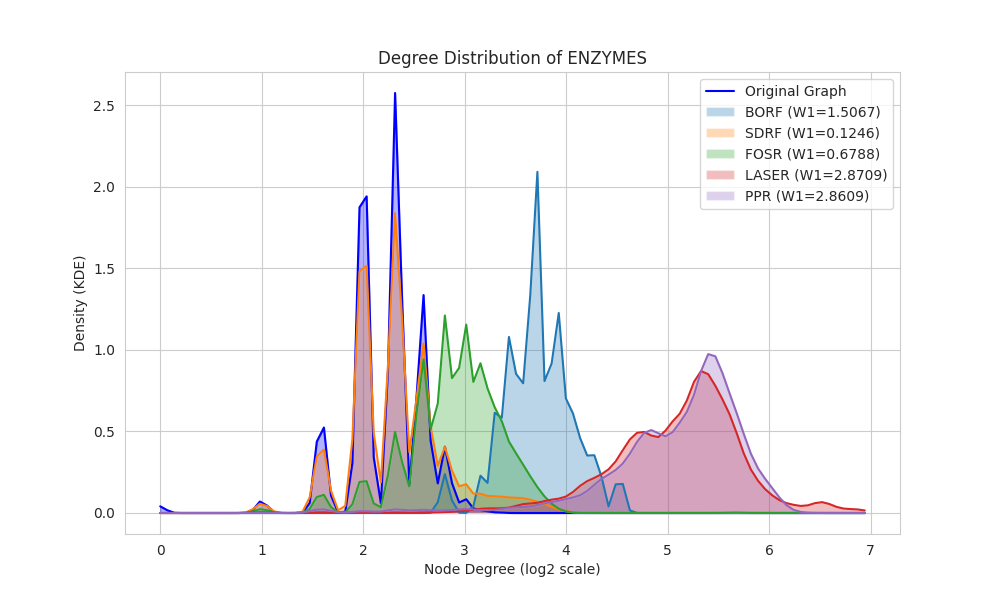}
                \caption{Original vs Rewired Degree Distribution for ENZYMES}
                \label{fig:URDDE}
            \end{subfigure}
            
            \caption{Comparing the concatenated Degree Distribution for all original (unrewired) and rewired graphs. The x-axis corresponds to the log-2 scale, the kernel density estimate of the Degree Distribution is used and underneath the Wassertein distance $W_1$ between the original and rewired graphs.}
            \label{fig:Comparisondegreedist}
        \end{figure}

\paragraph{Degree Distribution:} The Degree Distributions for all original and rewired graphs provides an overview of how rewiring changes graph structure across different datasets. Figure~\ref{fig:Comparisondegreedist} displays the Degree Distribution as an indicator of whether structural invariance is maintained. Those two metrics aim to understand which rewiring techniques preserve the graph topology most effectively. Table~\ref{tab:percentage_change_edgesaddedremoved} summarizes the number of edges added and removed by each rewiring technique.

From these plots, we aim to demonstrate that the closer the rewired graph's Degree Distribution density is to that of the original graph, the better the graph topology is conserved. \textbf{LASER} exhibits significant deviations from the original Degree Distribution across all datasets, as reflected by its higher W1 values. Specifically, \textbf{LASER} and \textbf{DIGL} consistently show larger shifts towards higher node degrees, indicating substantial topological alterations.

However, \textbf{SDRF} consistently maintains a Degree Distribution extremely close to the original graph across all datasets, evident from the very low W1 values. \textbf{BORF} and \textbf{FOSR} exhibit moderate shifts in comparison. Table~\ref{tab:percentage_change_edgesaddedremoved} indicates that most rewiring techniques primarily add edges to the graphs. This aligns with their primary objective of alleviating structural bottlenecks and mitigating over-squashing by enhancing connectivity within the graph.

Thus, Degree Distribution comparisons and W1 distances can represent crucial information regarding structural invariance.

    \begin{table}[h]
        \centering
        \caption{Percentage of edges added \%/removed \% by rewiring technique for each datasets of the original number of edges}
        \label{tab:percentage_change_edgesaddedremoved}
        \resizebox{\textwidth}{!}{
        \begin{tabular}{lcccccc}
            \toprule
            \textbf{Dataset} & \textbf{GTR} &\textbf{SDRF} & \textbf{FOSR} & \textbf{BORF} & \textbf{LASER} &\textbf{PPR} \\
            \midrule
            \textbf{IMBD-BINARY} &30.0\%/0.0\% &35.5\%/0.0\% &65.2\%/0.0\% &119.7\%/46.0\% &141.6\%/0.0\% &168.4\% / 0.0\% \\
            \textbf{MUTAG}   &110.7\%/0.0\% &4.6\%/0.0\% &271.8\%/0.0\%  &42.9\%/98.6\% & 668.3\%/0.0\% & 759.9\%/0.0\%     \\
            \textbf{ENZYMES} &38.9\%/0.0\% &14.8\%/0.0\% &92.9\%/0.0\%  & 476.8\%/54.62\%& 704.1\%/0.0\%& 733.5\%/0.0\% \\
            \textbf{PROTEINS}  &45.1\%/0.0\% &9.8\%/0.0\% &95.9\%/0.0\% &655.3\%/66.8\% &819.1\%/0.0\& &784.3\%/ 0.2\%  \\
            \bottomrule
        \end{tabular}
        }
    \end{table}

\paragraph{Similarity Metrics:} Figure ~\ref{fig:distance_metrics} in Appendix~\ref{appendix:similaritymetrics} displays the similarity metrics for the first, fourth, and last rewiring methods (ranked by performance) in three datasets. Generally, the lower performing rewiring methods are associated with greater distance metric values, possibly indicating that severe changes in structural properties is an undesirable result while rewiring. The Jaccard similarity is the only metric for which the higher performing rewiring methods consistently have greater values. As the Jaccard similarity is a measure of the difference in edge sets, this trend is explained by the addition of edges to reduce over-squashing.

For the Degree Distribution distance and the shortest path length distribution distance, the lowest performing rewiring method is the one that altered these distributions the most. The original degree and shortest path length distributions might then be characteristics that should ideally remain invariant when designing rewiring techniques.

\section{Discussion and Conclusion}
Our work introduces a framework, GRASP, for interpreting the role of structural invariance through rewiring for use in current graph-based architectures. Understanding which structural properties must remain invariant in effective rewiring techniques can be applied to constructing positional encodings for Graph Transformers, designing novel rewiring methods for GNNs, and supporting further work in graph topology research. To contribute to this understanding, we collect a variety of structural metrics before and after rewiring methods and compare them in the context of these methods' performances to reveal trends in structural invariance. We extend this analysis by visualizing changes in Degree Distribution and similarity metrics between the original and rewired data. 

This analysis of metrics can only suggest correlation and not strict causation, but nevertheless reveals trends that support the design of future rewiring techniques. Graph properties related to connectivity show a more consistent trend among higher performing rewiring techniques. Diameter and effective resistance are among the connectivity-related properties that consistently decrease for successful rewiring methods while the spectral gap and Forman curvature generally increase. Metrics that are more indicative of structural context than connectivity have less consistent trends across rewiring. Modularity and average betweenness centrality generally decrease while assortativity and clustering coefficient remain more similar to their original values. 

These patterns suggest curvature and bottleneck-related metrics are the most critical aspects to consider when rewiring for graph-based learning and that improving global communication takes priority over maintaining local community. Large improvements in connectivity-related metrics consistently resulted in better downstream performance whereas changes in strictly structural metrics varied. This is only true to some extent -  adding edges to create long-range connections cannot occur to the extent of rendering local information obsolete (Section~\ref{sub:sim}). The design of rewiring techniques should then strike a balance between adding enough edges to eliminate structural bottlenecks while maintaining the core community structure and connection tendencies. 

\paragraph{Limitation and Future Work}

In this work, we focused on a core set of structural metrics that balance informativeness and computational feasibility. While more complex measures such as graph edit distance---a robust but NP-hard similarity measure \cite{DABAH2021310}---and graphlet kernel distance---which captures subgraph frequency patterns \cite{shervashidze2009efficient}---were not included, our framework readily extends to incorporate such metrics. Expanding GRASP to a broader range of structural and distance measures, as well as larger datasets across diverse domains, represents a natural next step. Promising avenues include evaluating Cayley Graph Propagation \cite{wilson2024cayleygraphpropagation} within our framework, and leveraging recent insights that regular graph structures can enhance downstream GNN performance \cite{bechlerspeicher2024graphneuralnetworksuse}. Guiding graph regularity through our metrics could thus open up exciting directions for future research.


\section*{Author Contributions}

We, Catherine Aitken, Alexandre Benoit and He Yu of the University of Cambridge and Standford University, jointly declare that our work towards this project has been executed as follows:
\begin{itemize}
    \item {\bf Catherine Aitken} Conceptualization; Data curation; Formal analysis; Investigation; Methodology; Visualization; Writing – original draft; Writing – review \& editing.
    \item {\bf Alexandre Benoit} Conceptualization; Data curation; Formal analysis; Investigation; Methodology; Visualization; Writing – original draft; Writing – review \& editing.
    \item {\bf He Yu} Supervision; Project administration.
\end{itemize}

\section*{GitHub repository}

The companion source code for our project may be found at: \url{https://github.com/amgb20/L65-Mini-Project}.

\section*{Acknowledgements}

We thank our supervisor Yu He for guidance and discussion throughout the course of this work. We also thank Petar Veličković and Pietro Liò for the design and lectures of course L65. 

\clearpage
\printbibliography

\appendix
\section{APPENDIX}

\subsection{Rewiring methods}
\label{appendix:rewiring methods}

\paragraph{DiffWire \cite{diffwire} (2022) - differential-based.} DiffWire is a differentiable and inductive framework for graph rewiring in GNN. It rewires graphs by optimizing topological properties, specifically commute times and the spectral gap, through two layers integrated into the GNN architecture (Section~\ref{appendix:diff}). 

\paragraph{Stochastic Discrete Ricci Flow \cite{sdrf} (SDRF) (2022) - curvature-based.} SDRF utilizes curvature measures to alleviate over-squashing in GNNs. It introduces an edge-based curvature measure called the \textbf{Balanced Forman Curvature}, which offers a more computationally tractable alternative to classical measures like Ollivier curvature. Negatively curved edges contribute to graph bottlenecks, so SDRF selects these edges to rewire. 

\paragraph{Greedy Total Resistance \cite{gtr} (GTR) (2023) - effective resistance-based.} GTR minimizes the total effective resistance (Section~\ref{sec:strumetr}) of the graph using a greedy edge-adding heuristic. At each step, it adds the edge that would most reduce the total resistance. GTR optimizes this global measure of connectivity that describes the information flow across the entire graph.

\paragraph{Batch Ollivier-Ricci Flow \cite{borf} (BORF) (2023) - curvature-based.} BORF addresses over-smoothing and over-squashing by modifying the graph based on Ollivier-Ricci curvature. BORF takes edges with extreme curvature values and removes highly positively-curved edges (over-smoothing) and adds edges in negatively-curved regions (over-squashing) based on optimal transport plans between neighborhoods.

\paragraph{First-Order Spectral Rewiring \cite{fosr} (FOSR) (2023) - spectral-based.} FOSR strategically adds edges that have been predicted to maximize the spectral gap. At each step, it approximates the spectral gap (second eigenvector) of the normalized adjacency matrix and selects the edge whose addition is estimated to most improve connectivity by increasing the spectral gap.

\paragraph{Locality-Aware Sequential Rewiring \cite{laser} (LASER) (2024) - locally-based.} LASER is a spatial rewiring method that preserves both locality and sparsity in graphs. LASER incrementally adds edges over a sequence of rewiring "snapshots", which connect nodes that are n-hops apart. It looks for pairs of nodes that have weak connectivity by measuring how few short walks exist between those nodes and adds edges to the most poorly connected ones.

\subsection{Metrics equations}
\label{appendix:metrics equations}

\subsubsection{Structural Metrics}

\paragraph{Diameter - Connectivity-Related and Spatial:} This is the longest "short path" between any two nodes in a graph and reveals information about the graph's general reachability between nodes \cite{doi:10.1073/pnas.252631999}. 

\begin{equation}
    \text{diameter}(G) = \max \{ d(u, v) \mid u, v \text{ are in a connected component of } G \}
\end{equation}

\paragraph{Effective graph resistance - Connectivity-Related and Spectral:} This quantifies how efficiently information can flow through the graph by summing the effective resistances between all pairs of nodes. It uses electrical network theory to quantify how "resistant" the graph is to the information flow between two nodes \cite{ELLENS20112491}, and so is chosen as a metric for connectivity. In Equation~\ref{eq:egr}, $v$ represents the potential of a vertex, $I$ is the current passing through the vertices, and $R$ is the resistance.  
    \begin{equation}
        R_{ab} = \frac{v_a - v_b}{I}, \quad R_G = \sum_{1 \leq i < j \leq N} R_{ij}
        \label{eq:egr}
    \end{equation}

\paragraph{Modularity - Structural Context-Related and Spatial:} This measures the strength of community structure in a graph, reflecting the tendency for nodes within the same community to be densely connected, while sparsely connected to other communities. \cite{Clauset_2004} \cite{doi:10.1073/pnas.0601602103}. In Equation~\ref{eq:mod}, $n$ is the number of communities, $m$ is the total number of edges, $L_c$ is intra-community edges, $k_c$ is the sum of degrees of nodes in a community, and $\gamma$ is the resolution parameter. 
    \begin{equation}
        Q = \sum_{c=1}^{n} \left[ \frac{L_c}{m} - \gamma \left(\frac{k_c}{2m}\right)^2 \right]
        \label{eq:mod}
    \end{equation}

\paragraph{Degree Assortativity - Structural Context-Related and Spatial:} This quantifies the tendency of nodes to connect with others of similar degree. A positive value indicates assortative mixing (high-degree nodes linking to high-degree nodes), while a negative value indicates disassortative mixing \cite{Newman_2002}. We use it to capture structural tendencies via the Pearson correlation of node degrees, as shown in Equation~\ref{eq:da}, where $e_{xy}$ is the joint Degree Distribution, and $a_x$, $b_y$ are the corresponding marginals \cite{Newman_2003}. 

    \begin{equation}
        r = \frac{\sum_{xy} xy \left( e_{xy} - a_x b_y \right)}{\sigma_a \sigma_b}
        \label{eq:da}
    \end{equation}
    
\paragraph{Global Clustering Coefficient - Structural Context-Related and Spatial:} This measures the tendency of nodes in a graph to form clusters. An average clustering coefficient of 1 indicates a network that is composed entirely of cliques while a clustering coefficient of 0 indicates a network with no cliques, possibly indicating a bottleneck \cite{10.1093/oso/9780198805090.003.0007}. $T(u)$ is the number of triangles through node $u$ and $deg(u)$ is its degree. $C$ is the global clustering coefficient, taken as the average of all nodes' clustering coefficients. 
    \begin{equation}
        C = \frac{1}{n} \sum_{v \in G} c_v, \quad c_u = \frac{2T(u)}{\deg(u) (\deg(u) - 1)}
    \end{equation}

\paragraph{Spectral Gap - Connectivity-Related and Spectral:} It is taken as the smallest positive eigenvalue of the Laplacian matrix of the graph \cite{10.1093/imrn/rnz077}, which for a connected graph is the second eigenvalue $\lambda_2$. A larger spectral gap is indicative of stronger connectivity.

\paragraph{Forman Curvature - Connectivity-Related and Spatial:} This captures how paths within a graph diverge or converge and measures geodesic dispersion. It indicates whether two paths starting from nearby nodes stay parallel (zero curvature), converge (positive curvature), or diverge (negative curvature). Strongly negative curvature often signals the presence of structural bottlenecks, which contribute to over-squashing in GNNs \cite{topping2022understanding}. We choose Forman curvature \cite{forman} as our curvature metric due to its computational efficiency. Unlike Ollivier curvature, which is more computationally intensive and less localized, Forman curvature enables finer control over local structural properties. We calculate it using the \texttt{GraphRicciCurvature} \cite{ni2019community} Python library. For $v_1, v_2$ (two nodes) in a graph and $e$ (edge) between them, the general \textit{1D Forman curvature} of $e$ is given by (\cite{Sreejith_2016}):

\begin{equation}
F_{\text{full}}(e) = w_e \left( \frac{w_{v_1}}{w_e} + \frac{w_{v_2}}{w_e} 
- \sum_{e_{v_1} \sim e, e_{v_2} \sim e} 
\left[ \frac{w_{v_1}}{\sqrt{w_e w_{e_{v_1}}}} + \frac{w_{v_2}}{\sqrt{w_e w_{e_{v_2}}}} \right] 
\right),
\end{equation}
    
with $e_{v_1} \sim e$ and $e_{v_2} \sim e$: the edges other than $e$ that are adjacent to nodes $v_1$ and $v_2$ respectively; $w_e$, $w_{e_{v_1}}$, and $w_{e_{v_2}}$: the weights of $e$, $e_{v_1}$, and $e_{v_2}$ respectively; and $w_{v_1}$ and $w_{v_2}$:  weights of the nodes $v_1$ and $v_2$ respectively \cite{bober2022rewiringnetworksgraphneural}.

\paragraph{Average Betweenness Centrality - Structural Context-Related and Spatial:} This reflects how often a node lies on the shortest path between two other nodes, averaged across the network \cite{10.1093/oso/9780198805090.003.0007}. Lower values indicate that few nodes dominate path routing (e.g., over-squashing), while higher values imply more distributed connectivity.Equation~\ref{eq:abc} holds $\sigma(s,t)$ as the number of shortest $(s,t)$ paths and $\sigma(s,t|v)$, the number of those paths passing through $v$.
    \begin{equation}
        C = \frac{1}{|G|} \sum_{n \in G} c_n, \quad c_B(v) = \sum_{s, t \in V} \frac{\sigma(s, t \mid v)}{\sigma(s, t)}
        \label{eq:abc}
    \end{equation}

\subsubsection{Similarity Metrics Collection}
\paragraph{Jaccard similarity for edge sets:} This measures the similarity between edge sets of the original and rewired graph by finding the ratio of the size of the intersection between both edge sets to the size of their union \cite{9926326}. For any two finite sets A and B:  

\begin{equation}
    J(A, B) = \frac{|A \cap B|}{|A \cup B|} = \frac{|A \cap B|}{|A| + |B| - |A \cap B|},
\end{equation}

\paragraph{Laplacian spectrum distance:} Measures the p-norm of the difference between the sorted eigenvalue vectors of the original ($G$) and rewired graph ($G'$) comparing their spectral properties using Laplacian matrices \cite{PATANE201968}:

\begin{equation}
    d_{Lap}(G, G') = (\sum_{i=1}^n |\lambda_i - \lambda_i'|^p)^{1/p}
\end{equation}

with $\lambda$ and $\lambda_i'$ being the eigenvalues of $G$ and $G'$ sorted in ascending order and $p\in \mathbb{R^+}$.

\paragraph{Spectral Norm of Adjacency Difference:} Evaluates the largest single value of the difference between the adjacency $A$ and $A'$ matrices of $G$ and $G'$ \cite{gervens2022graphsimilaritybasedmatrix} as the spectral norm ($||\cdot||_2$):

\begin{equation}
    d_{Adj}(G, G') = || A - A'||_2
\end{equation}

\paragraph{Degree Distribution Difference:} Measures the Waaserstein (W1) distance between the Degree Distribution ($P_{G}$ and $Q_{G'}$ of $G$ and $G'$). This is, informally, the effort required to reconfigure one distribution into another \cite{Panaretos_2019}; the Degree Distribution in this context. 

\begin{equation}
    d_{\text{Deg}}(G, G') = W_1(P, Q) = \inf_{\gamma \in \Gamma(P, Q)} \mathbb{E}_{(x, y) \sim \gamma} [|x - y|]
\end{equation}

with $ \Gamma(P_G, Q_{G'})$ the set of all joint distributions with marginals $P_G$ and $Q_{G'}$.

\paragraph{Shortest Path Length Distribution Difference:} Calculates the W1 distance between the distribution of shortest path lengths in $G$ and $G'$ \cite{10.1007/978-3-642-17316-5_32}.
\subsection{Forman Curvature}
\label{appendix:curvature}

\begin{figure}[h]
    \centering
    \includegraphics[width=\textwidth]{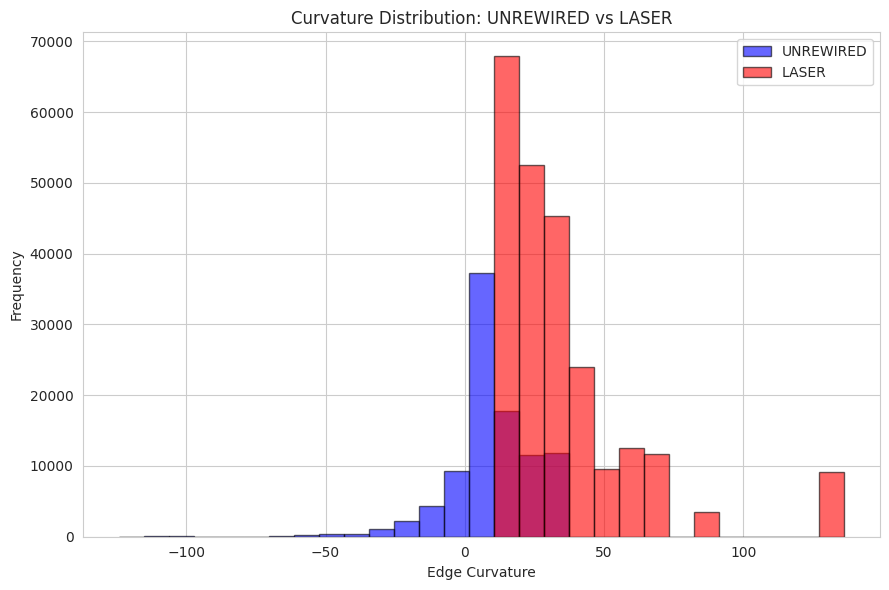}
    \caption{Original and LASER Forman Curvature Distribution}
    \label{fig:curvature_dist}
\end{figure}

\subsection{Similarity Metrics}
\label{appendix:similaritymetrics}

    \begin{figure}[h]
    \centering
    \begin{subfigure}{0.45\textwidth}
        \centering
        \includegraphics[width=\linewidth]{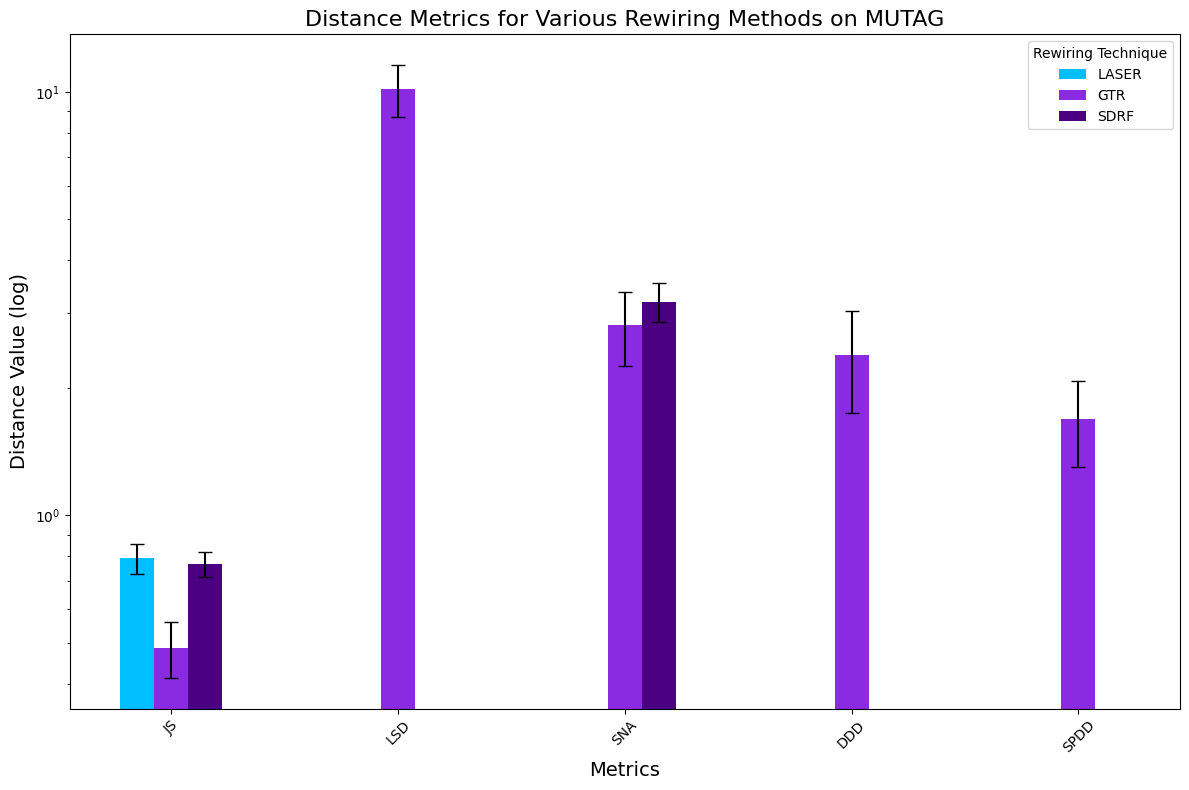}
        \caption{MUTAG Similarity Metrics}
    \end{subfigure}%
    \hspace{0.05\textwidth} 
    \begin{subfigure}{0.45\textwidth}
        \centering
        \includegraphics[width=\linewidth]{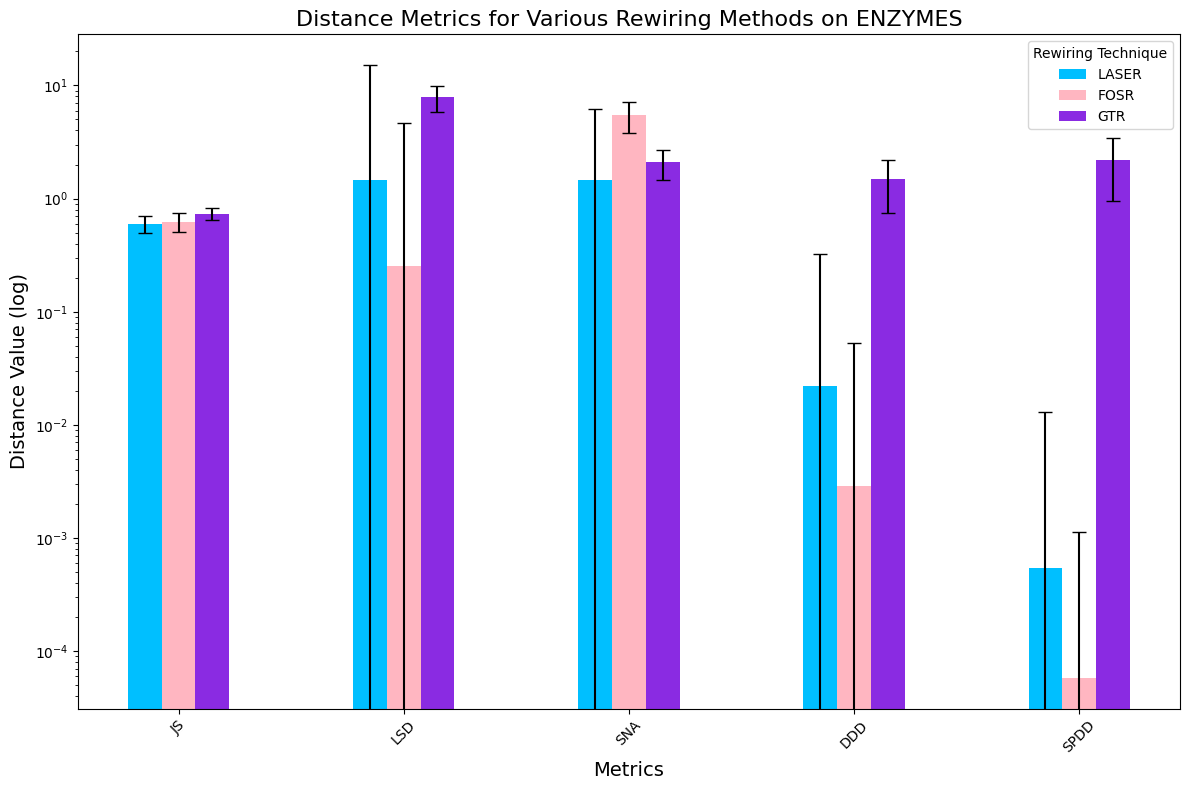}
        \caption{ENZYMES Similarity Metrics}
    \end{subfigure}

    \vspace{0.5cm} 
    \begin{subfigure}{0.5\textwidth}
        \centering
        \includegraphics[width=\linewidth]{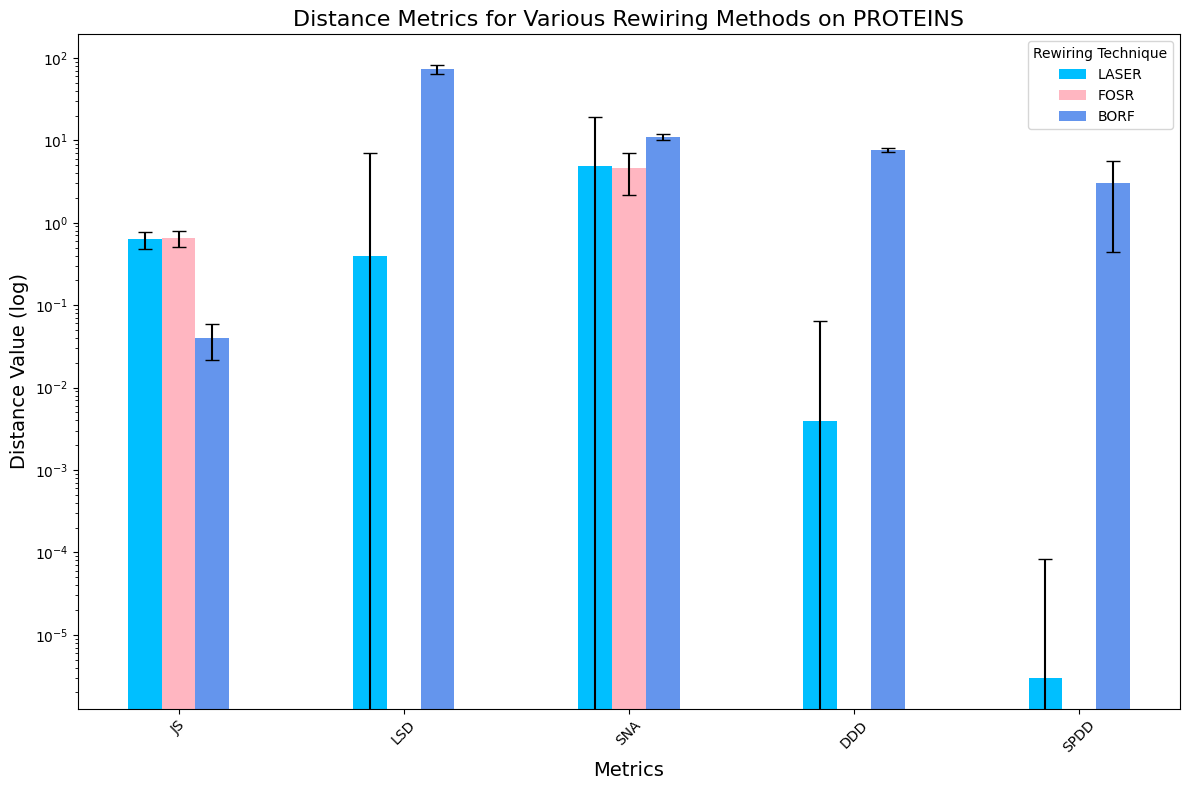}
        \caption{PROTEINS Similarity Metrics}
    \end{subfigure}
    \caption{Similarity metrics for MUTAG, ENZYMES, and PROTEINS across various rewiring techniques. Metrics from left to right are the Jaccard similarity, laplacian spectrum distance, spectral norm of adjacency difference, Degree Distribution distance, and shortest path length distribution difference.}
    \label{fig:distance_metrics}
\end{figure}

\subsection{EXPERIMENTAL SETUP}
    \subsubsection{Diffwire Rewiring Specifications}
    \label{appendix:diff}

    Diffwire comprises of two layers. The first, called \textbf{CT-LAYER}, learns commute times (related to effective resistance) and uses them to re-weight edges. The second, \textbf{GAP-LAYER}, directly optimizes the graph's spectral gap to enhance connectivity. Both layers are parameter-free and differentiable, enabling the model to adaptively predict optimal rewired topologies for unseen test graphs.

    Diffwire is implemented wholly within Google Colab for the purposes of this work.

    \subsubsection{Hardware Specification and Libraries}

    All experiments were conducted with Python 3.10.12 using PyTorch \cite{pytorch}, Numpy \cite{numpy}, Panda \cite{pandas}, with figures generated using Matplotlib \cite{matplotlib}.

    Data was retrieved from Google Sheets via Google Cloud integration.

    The experiments were run on a combination of local and cloud-based hardware. Local tests were executed on a 64-bit Intel(R) Core(TM) i7-9750H CPU @ 2.60GHz, while compute-intensive tasks were carried out on Google Colab using GPU and TPU instances, specifically the NVIDIA Tesla K80 with 12GB of VRAM.
    
\clearpage
\subsection{COMPLETE METRIC TABLES}
\begin{table}[H]
    \centering
    \caption{GTR, BORF, and FOSR on IMDB-BINARY}
    \label{tab:rewiring1}
    \resizebox{\textwidth}{!}{
    \begin{tabular}{lcccc}
        \toprule
        & \textbf{Original} & \textbf{GTR} & \textbf{BORF} & \textbf{FOSR} \\
        \midrule
        Diameter & 1.861 $\pm$ 0.346 & 1.832 $\pm$ 0.374 & 2.000 $\pm$ 0.000 & 1.557 $\pm$ 0.497 \\
        Effective Resistance & 0.285 $\pm$ 0.156 & 0.195 $\pm$ 0.066 & 0.196 $\pm$ 0.000 & 0.155 $\pm$ 0.046 \\
        Modularity & 0.298 $\pm$ 0.162 & 0.151 $\pm$ 0.134 & 0.074 $\pm$ 0.000 & 0.073 $\pm$ 0.098 \\
        Assortativity & -0.135 $\pm$ 0.163 & -0.081 $\pm$ 0.101 & -0.336 $\pm$ 0.000 & -0.056 $\pm$ 0.076 \\
        Clustering Coefficient & 0.947 $\pm$ 0.033 & 0.746 $\pm$ 0.157 & 0.522 $\pm$ 0.000 & 0.839 $\pm$ 0.181 \\
        Spectral Gap & 0.343 $\pm$ 0.311 & 0.664 $\pm$ 0.274 & 0.741 $\pm$ 0.000 & 0.849 $\pm$ 0.255 \\
        Forman Curvature & -15.529 $\pm$ 11.907 & -19.461 $\pm$ 10.746 & 0.025 $\pm$ 0.000 & 0.009 $\pm$ 0.010 \\
        Average Betweenness Centrality & 0.030 $\pm$ 0.016 & 0.052 $\pm$ 0.002 & 0.000 $\pm$ 0.000 & 0.000 $\pm$ 0.000 \\
        \bottomrule
    \end{tabular}
    }
\end{table}

\begin{table}[H]
    \centering
    \caption{DiffWire, SDRF, and LASER on IMDB-BINARY}
    \label{tab:rewiring2}
    \resizebox{\textwidth}{!}{
    \begin{tabular}{lcccc}
        \toprule
        & \textbf{Original} & \textbf{Diffwire} & \textbf{SDRF} & \textbf{LASER} \\
        \midrule
        Diameter & 1.861 $\pm$ 0.346 & 1.751 $\pm$ 0.638 & 1.822 $\pm$ 0.383 & 1.000 $\pm$ 0.000 \\
        Effective Resistance & 0.285 $\pm$ 0.156 & 1296.261 $\pm$ 6724.263 & 0.192 $\pm$ 0.078 & 0.117 $\pm$ 0.037 \\
        Modularity & 0.298 $\pm$ 0.162 & 0.024 $\pm$ 0.058 & 0.167 $\pm$ 0.129 & 0.000 $\pm$ 0.000 \\
        Assortativity & -0.135 $\pm$ 0.163 & -0.800 $\pm$ 0.261 & -0.127 $\pm$ 0.108 & NAN $\pm$ NAN \\
        Clustering Coefficient & 0.947 $\pm$ 0.033 & 0.255 $\pm$ 0.326 & 0.849 $\pm$ 0.129 & 1.000 $\pm$ 0.000 \\
        Spectral Gap & 0.343 $\pm$ 0.311 & 0.757 $\pm$ 0.323 & 0.545 $\pm$ 0.320 & 1.063 $\pm$ 0.021 \\
        Forman Curvature & -15.529 $\pm$ 11.907 & -15.624 $\pm$ 11.806 & 0.022 $\pm$ 0.015 & 0.000 $\pm$ 0.000 \\
        Average Betweenness Centrality & 0.030 $\pm$ 0.016 & 0.045 $\pm$ 0.025 & 0.019 $\pm$ 0.012 & 0.000 $\pm$ 0.000 \\
        \bottomrule
    \end{tabular}
    }
\end{table}

\begin{table}[H]
    \centering
    \caption{GTR, BORF, and FOSR on MUTAG}
    \label{tab:mutag_rewiring1}
    \resizebox{\textwidth}{!}{
    \begin{tabular}{lcccc}
        \toprule
        & \textbf{Original} & \textbf{GTR} & \textbf{BORF} & \textbf{FOSR} \\
        \midrule
        Diameter & 8.218 $\pm$ 1.842 & 2.819 $\pm$ 0.574 & 3.000 $\pm$ 0.000 & 2.426 $\pm$ 0.646 \\
        Effective Resistance & 0.850 $\pm$ 0.060 & 0.423 $\pm$ 0.061 & 0.000 $\pm$ 0.000 & nan $\pm$ nan \\
        Modularity & 0.464 $\pm$ 0.060 & 0.181 $\pm$ 0.066 & 0.180 $\pm$ 0.000 & 0.119 $\pm$ 0.074 \\
        Assortativity & -0.279 $\pm$ 0.169 & -0.151 $\pm$ 0.137 & -0.820 $\pm$ 0.000 & -0.131 $\pm$ 0.090 \\
        Clustering Coefficient & 0.000 $\pm$ 0.000 & 0.061 $\pm$ 0.107 & 0.000 $\pm$ 0.000 & 0.499 $\pm$ 0.285 \\
        Spectral Gap & 0.075 $\pm$ 0.031 & 0.558 $\pm$ 0.121 & 0.345 $\pm$ 0.000 & 0.670 $\pm$ 0.193 \\
        Forman Curvature & 0.005 $\pm$ 0.263 & -5.117 $\pm$ 1.27 & 0.135 $\pm$ 0.000 & 0.028 $\pm$ 0.012 \\
        Average Betweenness Centrality & 0.169 $\pm$ 0.023 & 0.052 $\pm$ 0.002 & 0.000 $\pm$ 0.000 & 0.000 $\pm$ 0.000 \\
        \bottomrule
    \end{tabular}
    }
\end{table}

\begin{table}[H]
    \centering
    \caption{Diffwire, SDRF, and LASER on MUTAG}
    \label{tab:mutag_rewiring2}
    \resizebox{\textwidth}{!}{
    \begin{tabular}{lcccc}
        \toprule
        & \textbf{Original} & \textbf{Diffwire} & \textbf{SDRF} & \textbf{LASER} \\
        \midrule
        Diameter & 8.218 $\pm$ 1.842 & 5.154 $\pm$ 2.094 & 8.027 $\pm$ 1.819 & 1.000 $\pm$ 0.000 \\
        Effective Resistance & 0.850 $\pm$ 0.060 & 86.161 $\pm$ 93.724 & nan $\pm$ nan & 0.000 $\pm$ 0.000 \\
        Modularity & 0.464 $\pm$ 0.060 & 0.616 $\pm$ 0.078 & 0.469 $\pm$ 0.053 & 0.000 $\pm$ 0.000 \\
        Assortativity & -0.279 $\pm$ 0.169 & -0.346 $\pm$ 0.198 & -0.260 $\pm$ 0.133 & NAN $\pm$ NAN \\
        Clustering Coefficient & 0.000 $\pm$ 0.000 & 0.000 $\pm$ 0.000 & 0.079 $\pm$ 0.040 & 1.000 $\pm$ 0.000 \\
        Spectral Gap & 0.075 $\pm$ 0.031 & 0.001 $\pm$ 0.006 & 0.075 $\pm$ 0.033 & 1.064 $\pm$ 0.019 \\
        Forman Curvature & 0.005 $\pm$ 0.263 & 0.846 $\pm$ 0.724 & 0.163 $\pm$ 0.022 & 0.000 $\pm$ 0.000 \\
        Average Betweenness Centrality & 0.169 $\pm$ 0.023 & 0.045 $\pm$ 0.057 & 0.000 $\pm$ 0.000 & 0.000 $\pm$ 0.000 \\
        \bottomrule
    \end{tabular}
    }
\end{table}

\begin{table}[H]
    \centering
    \caption{GTR, BORF, and FOSR on ENZYMES}
    \label{tab:enzymes_rewiring1}
    \resizebox{\textwidth}{!}{
    \begin{tabular}{lcccc}
        \toprule
        & \textbf{Original} & \textbf{GTR} & \textbf{BORF} & \textbf{FOSR} \\
        \midrule
        Diameter & 10.902 $\pm$ 4.828 & 3.535 $\pm$ 1.147 & 3.000 $\pm$ 0.000 & 3.295 $\pm$ 1.153 \\
        Effective Resistance & 1.063 $\pm$ 1.326 & 0.411 $\pm$ 0.181 & 0.117 $\pm$ 0.000 & 0.234 $\pm$ 0.116 \\
        Modularity & 0.571 $\pm$ 0.113 & 0.35 $\pm$ 0.141 & 0.143 $\pm$ 0.000 & -0.031 $\pm$ 0.112 \\
        Assortativity & -0.001 $\pm$ 0.172 & -0.007 $\pm$ 0.134 & -0.058 $\pm$ 0.000 & 0.320 $\pm$ 0.248 \\
        Clustering Coefficient & 0.453 $\pm$ 0.198 & 0.257 $\pm$ 0.161 & 0.336 $\pm$ 0.000 & 0.468 $\pm$ 0.249 \\
        Spectral Gap & 0.046 $\pm$ 0.133 & 0.356 $\pm$ 0.227 & 0.600 $\pm$ 0.000 & 0.027 $\pm$ 0.008 \\
        Forman Curvature & -3.072 $\pm$ 1.260 & -6.593 $\pm$ 1.955 & 0.019 $\pm$ 0.000 & 0.320 $\pm$ 0.142 \\
        Average Betweenness Centrality & 0.116 $\pm$ 0.040 & 0.038 $\pm$ 0.009 & 0.000 $\pm$ 0.000 & 0.027 $\pm$ 0.008 \\
        \bottomrule
    \end{tabular}
    }
\end{table}

\begin{table}[H]
    \centering
    \caption{Diffwire, SDRF, and LASER on ENZYMES}
    \label{tab:enzymes_rewiring2}
    \resizebox{\textwidth}{!}{
    \begin{tabular}{lcccc}
        \toprule
        & \textbf{Original} & \textbf{Diffwire} & \textbf{SDRF} & \textbf{LASER} \\
        \midrule
        Diameter & 10.902 $\pm$ 4.828 & 9.853 $\pm$ 4.361 & 10.355 $\pm$ 4.921 & 1.098 $\pm$ 0.304 \\
        Effective Resistance & 1.063 $\pm$ 1.326 & $(4.86 \times 10^3) \pm (3.40 \times 10^4)$ & 0.919 $\pm$ 1.326 & 0.084 $\pm$ 0.069 \\
        Modularity & 0.571 $\pm$ 0.113 & 0.579 $\pm$ 0.113 & 0.540 $\pm$ 0.138 & 0.019 $\pm$ 0.090 \\
        Assortativity & -0.001 $\pm$ 0.172 & -0.146 $\pm$ 0.214 & 0.118 $\pm$ 0.271 & 0.181 $\pm$ 0.498 \\
        Clustering Coefficient & 0.453 $\pm$ 0.198 & 0.344 $\pm$ 0.179 & 0.505 $\pm$ 0.183 & 0.993 $\pm$ 0.050 \\
        Spectral Gap & 0.046 $\pm$ 0.133 & 0.021 $\pm$ 0.072 & 0.054 $\pm$ 0.152 & 0.982 $\pm$ 0.223 \\
        Forman Curvature & -3.072 $\pm$ 1.260 & -2.670 $\pm$ 1.384 & 0.109 $\pm$ 0.043 & 0.000 $\pm$ 0.000 \\
        Average Betweenness Centrality & 0.116 $\pm$ 0.040 & 0.093 $\pm$ 0.051 & 0.000 $\pm$ 0.000 & 0.000 $\pm$ 0.001 \\
        \bottomrule
    \end{tabular}
    }
\end{table}

\begin{table}[H]
    \centering
    \caption{GTR, BORF, and FOSR on PROTEINS}
    \label{tab:proteins_rewiring1}
    \resizebox{\textwidth}{!}{
    \begin{tabular}{lcccc}
        \toprule
        & \textbf{Original} & \textbf{GTR} & \textbf{BORF} & \textbf{FOSR} \\
        \midrule
        Diameter & 11.571 $\pm$ 7.898 & 3.603 $\pm$ 1.986 & 2.527 $\pm$ 0.500 & 3.467 $\pm$ 2.453 \\
        Effective Resistance & 1.037 $\pm$ 1.336 & 0.411 $\pm$ 0.181 & 0.188 $\pm$ 0.008 & 0.233 $\pm$ 0.184 \\
        Modularity & 0.545 $\pm$ 0.185 & 0.327 $\pm$ 0.21 & 0.116 $\pm$ 0.040 & -0.021 $\pm$ 0.109 \\
        Assortativity & -0.065 $\pm$ 0.199 & -0.023 $\pm$ 0.104 & -0.200 $\pm$ 0.130 & 0.440 $\pm$ 0.330 \\
        Clustering Coefficient & 0.514 $\pm$ 0.231 & 0.353 $\pm$ 0.273 & 0.383 $\pm$ 0.131 & 0.545 $\pm$ 0.361 \\
        Spectral Gap & 0.096 $\pm$ 0.221 & 0.451 $\pm$ 0.346 & 0.677 $\pm$ 0.061 & 0.023 $\pm$ 0.012 \\
        Forman Curvature & -2.975 $\pm$ 1.175 & -6.898 $\pm$ 2.44 & 0.022 $\pm$ 0.003 & 0.329 $\pm$ 0.177 \\
        Average Betweenness Centrality & 0.118 $\pm$ 0.050 & 0.033 $\pm$ 0.014 & 0.000 $\pm$ 0.000 & 0.023 $\pm$ 0.012 \\
        \bottomrule
    \end{tabular}
    }
\end{table}

\begin{table}[H]
    \centering
    \caption{Diffwire, SDRF, and LASER on PROTEINS}
    \label{tab:proteins_rewiring2}
    \resizebox{\textwidth}{!}{
    \begin{tabular}{lcccc}
        \toprule
        & \textbf{Original} & \textbf{Diffwire} & \textbf{SDRF} & \textbf{LASER} \\
        \midrule
        Diameter & 11.571 $\pm$ 7.898 & 9.819 $\pm$ 7.170 & 11.150 $\pm$ 7.885 & 1.224 $\pm$ 0.466 \\
        Effective Resistance & 1.037 $\pm$ 1.336 & $(1.96 \times 10^3) \pm (1.82 \times 10^4)$ & 0.924 $\pm$ 1.301 & 0.109 $\pm$ 0.091 \\
        Modularity & 0.545 $\pm$ 0.185 & 0.561 $\pm$ 0.187 & 0.523 $\pm$ 0.200 & 0.032 $\pm$ 0.111 \\
        Assortativity & -0.065 $\pm$ 0.199 & -0.224 $\pm$ 0.282 & 0.033 $\pm$ 0.288 & 0.028 $\pm$ 0.337 \\
        Clustering Coefficient & 0.514 $\pm$ 0.231 & 0.359 $\pm$ 0.211 & 0.553 $\pm$ 0.217 & 0.988 $\pm$ 0.031 \\
        Spectral Gap & 0.096 $\pm$ 0.221 & 0.077 $\pm$ 0.216 & 0.113 $\pm$ 0.250 & 0.965 $\pm$ 0.252 \\
        Forman Curvature & -2.975 $\pm$ 1.175 & -2.573 $\pm$ 1.350 & 0.112 $\pm$ 0.052 & 0.000 $\pm$ 0.000 \\
        Average Betweenness Centrality & 0.118 $\pm$ 0.050 & 0.088 $\pm$ 0.059 & 0.000 $\pm$ 0.000 & 0.000 $\pm$ 0.001 \\
        \bottomrule
    \end{tabular}
    }
\end{table}

\subsection{COMPLETE PERCENTAGE CHANGE TABLES}

\begin{table}[H]
    \centering
    \caption{Percentage Change of Metrics for MUTAG with Various Rewiring Techniques}
    \label{tab:percentage_change_mutag}
    \resizebox{\textwidth}{!}{
    \begin{tabular}{lcccccc}
        \toprule
        & \textbf{DiffWire} & \textbf{GTR} & \textbf{SDRF} & \textbf{FOSR} & \textbf{BORF} & \textbf{LASER} \\
        \midrule
        Diameter & -37.28 & -65.70 & -2.33 & -70.49 & -63.49 & -87.33 \\
        Effective Resistance & 10036.59 & -50.24 & N/A & N/A & N/A & -85.93 \\
        Modularity & 32.76 & -60.99 & 1.15 & -74.44 & -61.28 & -100.00 \\
        Assortativity & 24.01 & -45.88 & -6.96 & -53.15 & 193.75 & N/A \\
        Clustering Coefficient & N/A & N/A & N/A & N/A & N/A & N/A \\
        Spectral Gap & -98.67 & 644.00 & 0.30 & 793.26 & 360.46 & 1318.57 \\
        Forman Curvature & 16820.00 & -102440.00 & 3157.70 & 453.04 & 2598.42 & -2135.32 \\
        Average Betweenness Centrality & -73.37 & -69.23 & -100.00 & -100.00 & -100.00 & -100.00 \\
        \bottomrule
    \end{tabular}
    }
\end{table}

\begin{table}[H]
    \centering
    \caption{Percentage Change of Metrics for ENZYMES with Various Rewiring Techniques}
    \label{tab:percentage_change_enzymes}
    \resizebox{\textwidth}{!}{
    \begin{tabular}{lcccccc}
        \toprule
        & \textbf{DiffWire} & \textbf{GTR} & \textbf{SDRF} & \textbf{FOSR} & \textbf{BORF} & \textbf{LASER} \\
        \midrule
        Diameter & -9.62 & -67.57 & -5.02 & -69.78 & -72.48 & -89.93\\
        Effective Resistance & $4.57 \times 10^5$ & -61.34 & -13.53 & -69.89 & -89.01 & -92.10\\
        Modularity & 1.40 & -38.70 & -5.35 & -58.93 & -74.90 & -96.70\\
        Assortativity & $1.45 \times 10^4$ & 600.00 & -11851.76 & 2974.97 & 5677.90 & -18169.90\\
        Clustering Coefficient & -24.06 & -43.27 & 11.42 & -29.42 & -25.86 & 119.24 \\
        Spectral Gap & -54.35 & 673.91 & 17.78 & 917.21 & 1204.42 & 2034.33\\
        Forman Curvature & -13.09 & 114.62 & -103.54 & -100.89 & -100.61 & -11337.07\\
        Average Betweenness Centrality & -19.83 & -67.24 & -100.00 & -100.00 & -100.00 & -99.88\\
        \bottomrule
    \end{tabular}
    }
\end{table}

\begin{table}[H]
    \centering
    \caption{Percentage Change of Metrics for IMDB-BINARY with Various Rewiring Techniques}
    \label{tab:percentage_change_imdb}
    \resizebox{\textwidth}{!}{
    \begin{tabular}{lcccccc}
        \toprule
        & \textbf{DiffWire} & \textbf{GTR} & \textbf{SDRF} & \textbf{FOSR} & \textbf{BORF} & \textbf{LASER} \\
        \midrule
        Diameter & -5.91 & -1.56 & -2.10 & -16.34 & 7.47 & -46.27 \\
        Effective Resistance & 454728.42 & -31.58 & -32.55 & -45.69 & -31.05 & -58.84 \\
        Modularity & -91.95 & -49.33 & -44.09 & -75.48 & -75.26 & -100.0 \\
        Assortativity & 492.59 & -40.00 & -6.05 & -58.59 & 149.15 & N/A \\
        Clustering Coefficient & -73.07 & -21.22 & -10.30 & -11.41 & -44.90 & 5.60 \\
        Spectral Gap & 120.70 & 93.59 & 58.86 & 147.56 & 116.10 & 209.83 \\
        Forman Curvature & 0.61 & 25.32 & 8.29 & 40.07 & -145.09 & 236.36 \\
        Average Betweenness Centrality & 50.00 & -36.67 & -28.23 & -69.63 & -18.13 & -100.00 \\
        \bottomrule
    \end{tabular}
    }
\end{table}

\begin{table}[H]
    \centering
    \caption{Percentage Change of Metrics for PROTEINS with Various Rewiring Techniques}
    \label{tab:percentage_change_proteins}
    \resizebox{\textwidth}{!}{
    \begin{tabular}{lcccccc}
        \toprule
        & \textbf{DiffWire} & \textbf{GTR} & \textbf{SDRF} & \textbf{FOSR} & \textbf{BORF} & \textbf{LASER}\\
        \midrule
        Diameter & -15.14 & -68.86 & -3.64 & -70.04 & -78.17 & -89.42\\
        Effective Resistance & $1.89 \times 10^5$ & -60.37 & -10.86 & -68.28 & -81.84 & -89.50\\
        Modularity & 2.94 & -40.00 & -3.99 & -57.29 & -78.70 & -94.12\\
        Assortativity & 244.62 & -64.62 & -151.15 & -68.38 & 207.44 & -142.84\\
        Clustering Coefficient & -30.16 & -31.32 & 7.50 & -14.41 & -25.42 & 92.13\\
        Spectral Gap & -19.79 & 369.79 & 17.64 & 468.22 & 604.76 & 904.98\\
        Forman Curvature & -13.51 & 131.87 & -103.76 & -100.77 & -100.72 & 12107.63\\
        Average Betweenness Centrality & -25.42 & -72.03 & -100.00 & -100.00 & -100.00 & -99.73\\
        \bottomrule
    \end{tabular}
    }
\end{table}

\subsubsection{Datasets}
\label{appendix:datasets}

\begin{table}[H]
    \centering
    \caption{Dataset characteristics including the number of graphs, classes, average nodes, and edges.}
    \begin{tabular}{lccccc}
        \hline
        \textbf{Dataset Name} & \textbf{Graphs} & \textbf{Classes} & \textbf{Avg. Nodes} & \textbf{Avg. Edges} \\
        \hline
        REDDIT-BINARY & 2000 & 2 & 429.63 & 497.75 \\
        IMDB-BINARY    & 1000 & 2 & 19.77  & 96.53  \\
        MUTAG         & 188 & 2 & 17.93  & 19.79  \\
        ENZYMES     & 600  & 6 & 32.63  & 62.14  \\
        PROTEINS     & 1113 & 2 & 39.06  & 72.82  \\
        COLLAB          & 5000 & 3 & 74.49  & 2457.78  \\
        \hline
    \end{tabular}
    \label{tab:dataset_characteristics}
\end{table}
\end{document}